\documentclass{sig-alternate-05-2015}

\usepackage{times}
\usepackage{epsfig}
\usepackage{graphicx}
\usepackage{amsmath}
\usepackage{amssymb}
\usepackage{algorithmic}

\def\sG{{\scriptscriptstyle {G}}}

\def\sT{{\scriptscriptstyle {T}}}

\def\sRm{{\scriptscriptstyle {R_m}}}
\def\sRm1{{\scriptscriptstyle {R_{m+1}}}}

\def\sCn{{\scriptscriptstyle {C}}}

\def\s1{{\scriptscriptstyle {-1}}}

\def\sCk1{{\scriptscriptstyle {C_{k+1}}}}

\def\I3{\mathbf I_{3}}

\newcommand*{\inlineequation}[2][]{%
  \begingroup
    \refstepcounter{equation}%
    \ifx\\#1\\%
    \else
      \label{#1}%
    \fi
    \relpenalty=10000 %
    \binoppenalty=10000 %
    \ensuremath{%
      #2%
    }%
  \endgroup
}
\usepackage[ruled,linesnumbered]{algorithm2e}
\newenvironment{Ualgorithm}[1][htpb]
  {\def\@algocf@post@ruled{\kern\interspacealgoruled\hrule  height\algoheightrule\kern3pt\relax}%
    \def\@algocf@capt@ruled{under}
    \begin{algorithm}[#1]}
  {\end{algorithm}}
  \DeclareMathOperator{\sign}{sign}
\usepackage[pagebackref=true,breaklinks=true,letterpaper=true,colorlinks,bookmarks=false]{hyperref}
\makeatletter
\DeclareRobustCommand\onedot{\futurelet\@let@token\@onedot}
\def\@onedot{\ifx\@let@token.\else.\null\fi\xspace}

\def\eg{\emph{e.g}\onedot} 
\def\ie{\emph{i.e}\onedot}

\def\etal{\emph{et al}\onedot}
\makeatother

\begin{document}

% Copyright
%\setcopyright{acmcopyright}
%\setcopyright{acmlicensed}
%\setcopyright{rightsretained}
%\setcopyright{usgov}
%\setcopyright{usgovmixed}
%\setcopyright{cagov}
%\setcopyright{cagovmixed}

% DOI 
%\nodoi{ }
%
%% ISBN
%\noisbn{ }
%
%%Conference
%\conferenceinfo{PLDI '13}{June 16--19, 2013, Seattle, WA, USA}
%
%\acmPrice{\$15.00}
%
%%
%% --- Author Metadata here ---
%\conferenceinfo{WOODSTOCK}{'97 El Paso, Texas USA}
%\CopyrightYear{2007} % Allows default copyright year (20XX) to be over-ridden - IF NEED BE.
%\crdata{0-12345-67-8/90/01}  % Allows default copyright data (0-89791-88-6/97/05) to be over-ridden - IF NEED BE.
% --- End of Author Metadata ---

\title{An Efficient Algebraic Solution to the Perspective-Three-Point Problem}


\numberofauthors{2} %  in this sample file, there are a *total*
\author{
% 1st. author
\alignauthor
Tong Ke\\
       \affaddr{University of Minnesota}\\
       \affaddr{Minneapolis, MN 55455}\\
       \email{kexxx069@cs.umn.edu}
% 2nd. author
\alignauthor
Stergios Roumeliotis\\
       \affaddr{University of Minnesota}\\
       \affaddr{Minneapolis, MN 55455}\\
       \email{stergios@cs.umn.edu}
}
% There's nothing stopping you putting the seventh, eighth, etc.
% author on the opening page (as the 'third row') but we ask,
% for aesthetic reasons that you place these 'additional authors'
% in the \additional authors block, viz.
%\additionalauthors{Additional authors: John Smith (The Th{\o}rv{\"a}ld Group,
%email: {\texttt{jsmith@affiliation.org}}) and Julius P.~Kumquat
%(The Kumquat Consortium, email: {\texttt{jpkumquat@consortium.net}}).}
% Just remember to make sure that the TOTAL number of authors
% is the number that will appear on the first page PLUS the
% number that will appear in the \additionalauthors section.

\maketitle
\begin{abstract}
In this work, we present an algebraic solution to the classical perspective-3-point (P3P) problem for determining the position and attitude of a camera from observations of three known reference points.
In contrast to previous approaches, we first directly determine the camera's attitude by employing the corresponding geometric constraints to formulate a system of trigonometric equations.
This is then efficiently solved, following an algebraic approach,  to determine the unknown rotation matrix and subsequently the camera's position.
As compared to recent alternatives, our method avoids computing unnecessary (and potentially numerically unstable) intermediate results, and thus achieves higher numerical accuracy and robustness at a lower computational cost.
These benefits are validated through extensive Monte-Carlo simulations for both nominal and close-to-singular geometric configurations.
\end{abstract}

%\begin{CCSXML}
%<ccs2012>
%  <concept_id>10010520.10010553.10010562</concept_id>
%  <concept_desc>Computer systems organization~Embedded systems</concept_desc>
%  <concept_significance>500</concept_significance>
%  <concept_id>10010520.10010575.10010755</concept_id>
%  <concept_desc>Computer systems organization~Redundancy</concept_desc>
%  <concept_significance>300</concept_significance>
%  <concept_id>10010520.10010553.10010554</concept_id>
%  <concept_desc>Computer systems organization~Robotics</concept_desc>
%  <concept_significance>100</concept_significance>
%  <concept_id>10003033.10003083.10003095</concept_id>
%  <concept_desc>Networks~Network reliability</concept_desc>
%  <concept_significance>100</concept_significance>
%</ccs2012>  
%\end{CCSXML}
%
%\ccsdesc[500]{Computer systems organization~Embedded systems}


%
%  Use this command to print the description
%
%\printccsdesc

% We no longer use \terms command

%\keywords{ACM proceedings; \LaTeX; text tagging}

%%%%%%%%%%%%%%%%%%%%%%%%%%%%%%%%%%%%%%%%%%%%
%
%
\section{Introduction} 
\label{sec:intro}
%
%
%%%%%%%%%%%%%%%%%%%%%%%%%%%%%%%%%%%%%%%%%%%%
%
%
%The Perspective-n-Point (PnP) problem aims at estimating the position and orientation of a calibrated camera with respect to a reference frame from $ n $ correspondences between 3D reference points and their 2D projections on an image. It is an old problem in computer vision and photogrammetry which is originated from camera calibration~\cite{fischler1981random,quan1999linear,abidi1995new,horaud1989analytic}. 
%
%The Perspective-n-Point (PnP) is a classical problem in computer vision and robotics, of which aim is estimating the position and orientation of a camera from observations of known features.
%%
%
The Perspective-n-Point (PnP) is the problem of determining the 3D position and orientation (pose) of  a camera from
observations of known point features.
The PnP is typically formulated and solved linearly by employing lifting (\eg,~\cite{ansar2003linear}), or as a nonlinear least-squares problem minimized iteratively (\eg,~\cite{haralick1989pose}) or directly (\eg,~\cite{hesch2011direct}).
The minimal case of the PnP (for n=3) is often used in practice, in conjunction with RANSAC, for removing outliers~\cite{fischler1981random}.
%
%In general, the PnP can be solved by either formulating and solving a linear system~\cite{ansar2003linear}, or optimizing a nonlinear least-squares cost function iteratively~\cite{haralick1989pose}, or directly~\cite{hesch2011direct}. 
%
%In 1981, Fischler and Bolles, when presenting the RANSAC algorithm~\cite{fischler1981random}, named this problem as the Perspective-n-Point (PnP), and a special case ($ n=3 $) as P3P. 
%%
%The P3P is the minimum subset of the PnP that has a finite number (up to four) of solutions, and hence it can be used to generate potential hypotheses and remove outliers when solving PnP by RANSAC.

The first solution to the P3P problem was given by Grunert~\cite{grunert1841pothenotische} in 1841. 
Since then, several methods have been introduced, some of which~\cite{grunert1841pothenotische,finsterwalder1903ruckwartseinschneiden,merritt1949explicit,fischler1981random,linnainmaa1988pose,grafarend1989dreidimensionaler} were reviewed and compared, in terms of numerical accuracy, by Haralick \etal~\cite{haralick1991analysis}. 
Common to these algorithms is that they employ the law of cosines to formulate a system of three quadratic equations in the features' distances from the camera. 
They differ, however, in the elimination process followed for arriving at a univariate polynomial.
Later on, Quan and Lan~\cite{quan1999linear}  and more recently Gao \etal~\cite{gao2003complete} employed the same formulation but instead used the Sylvester resultant~\cite{cox2006using} and Wu-Ritz's zero-decomposition method~\cite{wen1986basic}, respectively, to solve the resulting system of equations, and, in the case of~\cite{gao2003complete}, to determine the number of real solutions.
Regardless of the approach followed, once the feature's distances have been computed, finding the camera's orientation, expressed as a unit quaternion~\cite{horn1987closed} or a rotation matrix~\cite{horn1988closed}, often requires computing the eigenvectors of a $4\times4$~matrix (\eg,~\cite{quan1999linear}) or performing singular value decomposition (SVD) of a $3\times3$~matrix (\eg,~\cite{gao2003complete}), respectively, both of which are time-consuming. 
Furthermore, numerical error propagation from the computed distances to the rotation matrix significantly reduces the accuracy of the computed pose estimates.

To the best of our knowledge, the first method\footnote{Nister and Stewenius~\cite{nister2007minimal} also follow a geometric approach for solving the {\em generalized} P3P resulting into an octic univariate polynomial whose odd monomials vanish for the case of the central P3P.}
that does not employ the law of cosines in its P3P problem formulation is that of Kneip~\etal~\cite{kneip2011novel}, and later on that of Masselli and Zell~\cite{masselli2014new}.
Specifically,~\cite{kneip2011novel} and~\cite{masselli2014new}  follow a geometric approach for avoiding computing the features' distances and instead directly solve for the camera's pose.
In both cases, however, several intermediate terms (\eg,  tangents and cotangents of certain angles) need to be computed, which negatively affect the speed and numerical precision of the resulting algorithms.

%%%

%To the best of our knowledge, the first method that does not employ the law of cosines in its problem formulation is that of Nister and Stewenius~\cite{nister2007minimal}. 
%%
%Instead, they follow a geometric approach for solving the generalized (non-central) camera P3P resulting into an octic univariate polynomial whose odd monomials vanish for the case of the central P3P.
%%
%Later on, Kneip~\etal~\cite{kneip2011novel} and Masselli and Zell~\cite{masselli2014new} also introduced geometric methods for avoiding computing the features' distances and instead directly solved for the camera's pose.
%% 
%In both cases, however, several intermediate terms (\eg,  tangents and cotangents of certain angles) need to be computed, which negatively affect the speed and numerical precision of the resulting algorithms.

%%%

Similar to~\cite{kneip2011novel} and~\cite{masselli2014new}, our proposed approach does not require first computing the features' distances.
Differently though, in our derivation, we first eliminate the camera's position and the features' distances to result into a system of three equations involving {\em only the camera's orientation}.
Then, we follow an algebraic process for successively eliminating two of the unknown 3-dof and arriving into a quartic polynomial.
Our algorithm (summarized in Alg.~\ref{alg:p3p}) requires fewer operations and involves simpler and numerically more stable expressions, as compared to either~\cite{kneip2011novel} or~\cite{masselli2014new}, and thus performs better in terms of efficiency, accuracy, and robustness. 
%
%%
%
%Similar to~\cite{kneip2011novel,masselli2014new}, our method does not require computing the features' distances while a quartic polynomial is attained by changing variables. 
%%
%By following an alternative algebraic approach, however, that involves simpler and numerically more stable expressions as compared to~\cite{kneip2011novel,masselli2014new}, our algorithm performs better in terms of efficiency, accuracy, and robustness. 
%%
%
Specifically, the main advantages of our approach are:
\begin{itemize}
\item Our algorithm's implementation takes about 40\% of the time required by the current state of the art~\cite{kneip2011novel}.
\footnote{Although Masselli and Zell~\cite{masselli2014new} claim that their algorithm runs faster than Kneip \etal's~\cite{kneip2011novel}, our results (see Section~\ref{sec:results}) show the opposite to be true (by a small margin). 
The reason we arrive at a different conclusion is that our simulation randomly generates a new geometric configuration for each run, while Masselli employs only one configuration during their entire simulation, in which they save time due to caching.}
\item Our method achieves better accuracy than~\cite{kneip2011novel,masselli2014new} under nominal conditions.
Moreover, we are able to further improve the numerical precision by applying root polishing to the solutions of the quartic polynomial while remaining faster than~\cite{kneip2011novel,masselli2014new}.
%
%In addition, we achieve significant accuracy improvement by applying root polishing when solving the quartic equation while still remaining faster, without root polishing.
%
\item Our algorithm is more robust than~\cite{kneip2011novel,masselli2014new} when considering close-to-singular configurations (the three points are almost collinear or very close to each other).
\end{itemize}

The remaining of this paper is structured as follows. 
Section~\ref{sec:main} presents the definition of the P3P problem, as well as our derivations for estimating first the orientation and then the position of the camera. 
In Section~\ref{sec:results}, we assess the performance of our approach against ~\cite{kneip2011novel} and~\cite{masselli2014new}  in simulation for both nominal and singular configurations. 
Finally, we conclude our work in Section~\ref{sec:conclusion}.

\section{Problem Formulation and Solution} 
\label{sec:main}
%
%
%%%%%%%%%%%%%%%%%%%%%%%%%%%%%%%%%%%%%%%%%%%%%

\subsection{Problem Definition}
Given the positions, $ ^\sG\mathbf{p}_i $, of three known features $ f_i, \ i=1,2,3$, with respect to a reference frame $ \{G\} $, and the corresponding unit-vector, bearing measurements, $ ^\sCn\mathbf{b}_i$,  $i=1,2,3 $, our objective is to estimate the position, $ ^\sG\mathbf{p}_\sCn $, and orientation, \ie, the rotation matrix $ ^\sG_\sCn\mathbf{C} $, of the camera $ \{C\} $.

\subsection{Solving for the orientation}
\begin{figure}
\center
\includegraphics[width=.8\linewidth]{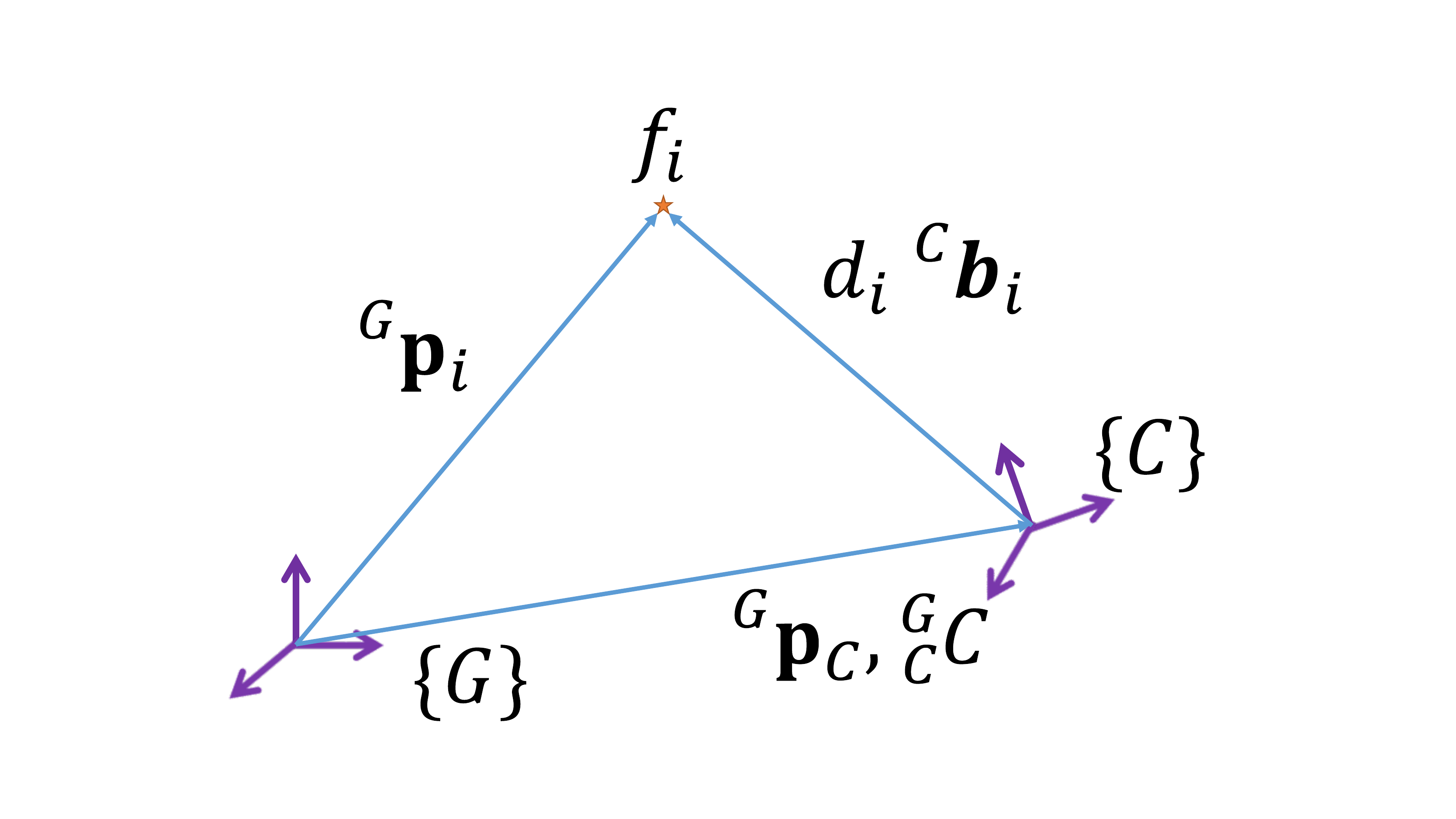}
\caption{The camera $ \{C\} $, whose position, $ {}^\sG\mathbf{p}_\sCn $, and orientation, $ {}^\sG_\sCn\mathbf{C} $, we seek to determine, observes unit-vector bearing measurement $ {}^\sCn\mathbf{b}_i $ of a feature $ f_i $, whose position, $ {}^\sG\mathbf{p}_i $, is known.}
\label{fig:di}
\end{figure}

From the geometry of the problem (see Fig.~\ref{fig:di}), we have (for $ i=1,2,3 $):\\
\begin{align} 
\label{eq:di} ^\sG\mathbf{p}_i&={}^\sG\mathbf{p}_\sCn+d_i{}^\sG_\sCn\mathbf{C}{}^\sCn\mathbf{b}_i
\end{align}
where $ d_i\triangleq\|^\sG\mathbf{p}_\sCn-{}^\sG\mathbf{p}_i\| $ is the distance between the camera and the feature $f_i$.\\
In order to eliminate the unknown camera position, $ {}^\sG\mathbf{p}_\sCn $, and feature distance, $ d_i,\ i=1,2,3 $, we subtract pairwise the three equations corresponding to \eqref{eq:di} for $ (i,j)=(1,2),\ (1,3)$ and $(2,3)$, and project them on the vector $ {}^\sG_\sCn\mathbf{C}({}^\sCn\mathbf{b}_i\times{}^\sCn\mathbf{b}_j) $ to yield the following system of 3 equations in the unknown rotation $ {}^\sG_\sCn\mathbf{C} $: 
\begin{align}
\label{eq:c1} (^\sG\mathbf{p}_1-{}^\sG\mathbf{p}_2)^\sT{}^\sG_\sCn\mathbf{C}({}^\sCn\mathbf{b}_1\times{}^\sCn\mathbf{b}_2)&=0\\
\label{eq:c2} (^\sG\mathbf{p}_1-{}^\sG\mathbf{p}_3)^\sT{}^\sG_\sCn\mathbf{C}({}^\sCn\mathbf{b}_1\times{}^\sCn\mathbf{b}_3)&=0\\
\label{eq:c3} (^\sG\mathbf{p}_2-{}^\sG\mathbf{p}_3)^\sT{}^\sG_\sCn\mathbf{C}({}^\sCn\mathbf{b}_2\times{}^\sCn\mathbf{b}_3)&=0
\end{align}
Next, and in order to compute one of the 3 unknown degrees of rotational freedom, we introduce the following factorization of ${}^\sG_\sCn\mathbf{C} $:
\begin{equation}
\label{eq:ccc}{}^\sG_\sCn\mathbf{C}=\mathbf{C}(\mathbf{k}_1,\theta_1)\mathbf{C}(\mathbf{k}_2,\theta_2)\mathbf{C}(\mathbf{k}_3,\theta_3)
\end{equation}
where\footnote{$ \mathbf{C}(\mathbf{k},\theta) $ denotes the rotation matrix describing the rotation about the unit vector, $ \mathbf{k}$, by an angle $ \theta $. Note that in the ensuing derivations, all rotation angles are defined using the left-hand rule.}\\
\begin{align}
\label{eq:ki}\mathbf{k}_1\triangleq\frac{{}^\sG\mathbf{p}_1-{}^\sG\mathbf{p}_2}{\|{}^\sG\mathbf{p}_1-{}^\sG\mathbf{p}_2\|},\ \mathbf{k}_3\triangleq\frac{{}^\sCn\mathbf{b}_1\times{}^\sCn\mathbf{b}_2}{\|{}^\sCn\mathbf{b}_1\times{}^\sCn\mathbf{b}_2\|},\ \mathbf{k}_2\triangleq\frac{\mathbf{k}_1\times\mathbf{k}_3}{\|\mathbf{k}_1\times\mathbf{k}_3\|}
\end{align}
Substituting \eqref{eq:ccc} in \eqref{eq:c1}, yields a scalar equation in the unknown $ \theta_2 $:
\begin{align}
\label{eq:t2}\mathbf{k}_1^\sT\mathbf{C}(\mathbf{k}_2,\theta_2)\mathbf{k}_3&=0
\end{align}
which we solve by employing Rodrigues' rotation formula~\cite{koks2006roundabout}:\footnote{$ \lfloor\mathbf{k}\rfloor $ denotes the $ 3\times3 $ skew-symmetric matrix corresponding to $ \mathbf{k} $ such that $ \lfloor\mathbf{k}\rfloor\mathbf{a}=\mathbf{k}\times\mathbf{a}$, $\forall\ \mathbf{k},\mathbf{a}\in\mathbb{R}^3$. Note also that if $ \mathbf{k} $ is a unit vector, then $ \lfloor\mathbf{k}\rfloor^2=\mathbf{k}\mathbf{k}^\sT-\mathbf{I} $, while for two vectors $ \mathbf{a}$, $\mathbf{b} $, $ \lfloor\mathbf{a}\rfloor\lfloor\mathbf{b}\rfloor=\mathbf{b}\mathbf{a}^\sT-(\mathbf{a}^\sT\mathbf{b})\mathbf{I} $. Lastly, it is easy to show that $ \lfloor\lfloor\mathbf{a}\rfloor\mathbf{b}\rfloor=\mathbf{b}\mathbf{a}^\sT-\mathbf{a}\mathbf{b}^\sT. $}
\begin{equation}
\label{eq:Rod}\mathbf{C}(\mathbf{k}_2,\theta_2)=\cos\theta_2\mathbf{I}-\sin\theta_2\lfloor\mathbf{k}_2\rfloor+(1-\cos\theta_2)\mathbf{k}_2\mathbf{k}_2^\sT
\end{equation}
to get
\begin{align}
\label{eq:st2}\theta_2=\arccos(\mathbf{k}_1^\sT\mathbf{k}_3)\pm\frac{\pi}{2}
\end{align}
Note that we only need to consider one of these two solutions [in our case, we select $ \theta_2=\arccos(\mathbf{k}_1^\sT\mathbf{k}_3)-\frac{\pi}{2} $; see Fig.~\ref{fig:ki}], since the other one will result in the same $ {}^\sG_\sCn\mathbf{C} $ (see Appendix~\ref{ssec:theta2} for a formal proof).\\
\indent In what follows, we describe the process for eliminating $ \theta_3 $ from \eqref{eq:c2} and \eqref{eq:c3}, and eventually arriving into a quartic polynomial involving a trigonometric function of $ \theta_1 $. 
To do so, we once again substitute in \eqref{eq:c2} and \eqref{eq:c3} the factorization of $ {}^\sG_\sCn\mathbf{C} $ defined in $\eqref{eq:ccc} $ to get (for $ i=1,2 $):
\begin{equation}
\label{eq:uCv}\mathbf{u}_i^\sT\mathbf{C}(\mathbf{k}_1,\theta_1)\mathbf{C}(\mathbf{k}_2,\theta_2)\mathbf{C}(\mathbf{k}_3,\theta_3)\mathbf{v}_i=0
\end{equation}
where
\begin{align}
\label{eq:ui}\mathbf{u}_i\triangleq{}^\sG\mathbf{p}_i-{}^\sG\mathbf{p}_3,~~
\mathbf{v}_i&\triangleq{}^\sCn\mathbf{b}_i\times{}^\sCn\mathbf{b}_3,~~i=1,2,
\end{align}
and employ the following property of rotation matrices
\begin{equation}
\mathbf{C}(\mathbf{k}_1,\theta_1)\mathbf{C}(\mathbf{k}_2,\theta_2)\mathbf{C}^\sT(\mathbf{k}_1,\theta_1)=\mathbf{C}(\mathbf{C}(\mathbf{k}_1,\theta_1)\mathbf{k}_2,\theta_2)\nonumber
\end{equation}
to rewrite \eqref{eq:uCv} in a simpler form as
\begin{align}
\mathbf{u}_i^\sT\mathbf{C}(\mathbf{k}_1,\theta_1)\mathbf{C}(\mathbf{C}(\mathbf{k}_2,\theta_2)\mathbf{k}_3,\theta_3)\mathbf{C}(\mathbf{k}_2,\theta_2)\mathbf{v}_i&=0\nonumber\\
\Rightarrow\mathbf{u}_i^\sT\mathbf{C}(\mathbf{k}_1,\theta_1)\mathbf{C}(\mathbf{k}^\prime_3,\theta_3)\mathbf{v}^\prime_i&=0\label{eq:k3prime}
\end{align}
where
\begin{align}
\mathbf{v}^\prime_i&\triangleq\mathbf{C}(\mathbf{k}_2,\theta_2)\mathbf{v}_i,\ i=1,2\nonumber\\
\label{eq:k2xk1}\mathbf{k}^\prime_3&\triangleq\mathbf{C}(\mathbf{k}_2,\theta_2)\mathbf{k}_3=\mathbf{k}_2\times\mathbf{k}_1 
\end{align}
The last equality in \eqref{eq:k2xk1} is geometrically depicted in Fig.~\ref{fig:ki} and algebraically derived in Appendix~\ref{ssec:k3p}. Analogously, it is straightforward to show that
\begin{equation}
\mathbf{k}^\prime_1\triangleq\mathbf{C}^\sT(\mathbf{k}_2,\theta_2)\mathbf{k}_1=\mathbf{k}_3\times\mathbf{k}_2\nonumber
\end{equation}
\begin{figure}
\center
\includegraphics[width=\linewidth]{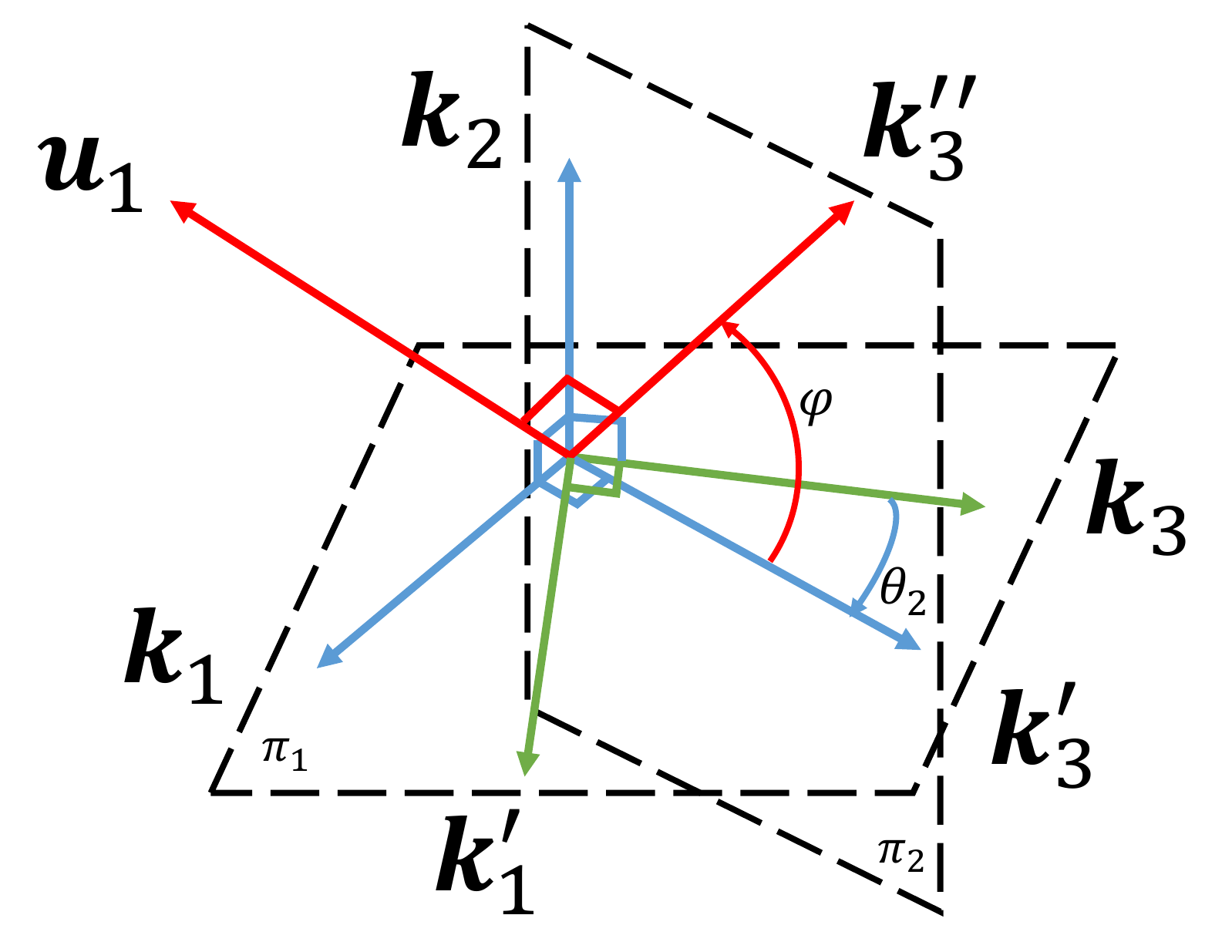}
\caption{Geometric relation between unit vectors $ \mathbf{k}_1,\ \mathbf{k}_2,\ \mathbf{k}_3,\ \mathbf{k}_3^\prime,\ \mathbf{k}_3^{\prime\prime},$ and $ \mathbf{u}_1 $. Note that $ \mathbf{k}_1,\ \mathbf{k}_3$, and $\mathbf{k}_3^\prime $ belong to a plane $ \pi_1 $ whose normal is $ \mathbf{k}_2 $. Also, $ \mathbf{k}_2,\ \mathbf{k}_3^\prime$, and $\mathbf{k}_3^{\prime\prime} $ lie on a plane, $ \pi_2 $, normal to $\pi_1$.}
\label{fig:ki}
\end{figure}
Next, by employing Rodrigues' rotation formula [see \eqref{eq:Rod}], for expressing the product of a rotation matrix and a vector as a linear function of the unknown $ \begin{bmatrix}\cos\theta & \sin\theta\end{bmatrix}^\sT  $, \ie,
\begin{align}
\mathbf{C}(\mathbf{k},\theta)\mathbf{v}&=(-\cos\theta\lfloor\mathbf{k}\rfloor^2-\sin\theta\lfloor\mathbf{k}\rfloor+\mathbf{k}\mathbf{k}^\sT)\mathbf{v}\nonumber\\
&=\begin{bmatrix}
-\lfloor\mathbf{k}\rfloor^2\mathbf{v} & -\lfloor\mathbf{k}\rfloor \mathbf{v}
\end{bmatrix}
\begin{bmatrix}
\cos\theta\\
\sin\theta
\end{bmatrix}+(\mathbf{k}^\sT\mathbf{v})\mathbf{k}\label{eq:Cv}
\end{align}
in \eqref{eq:k3prime} yields (for $ i=1,2 $):
\begin{align}
&\left(\begin{bmatrix}
-\lfloor\mathbf{k}_1\rfloor^2\mathbf{u}_i & \lfloor\mathbf{k}_1\rfloor \mathbf{u}_i
\end{bmatrix}
\begin{bmatrix}
\cos\theta_1\\
\sin\theta_1
\end{bmatrix}+(\mathbf{k}_1^\sT\mathbf{u}_i)\mathbf{k}_1\right)^\sT\nonumber\\
\cdot&\left(\begin{bmatrix}
-\lfloor\mathbf{k}^{\prime}_3\rfloor^2\mathbf{v}^{\prime}_i & -\lfloor\mathbf{k}^{\prime}_3\rfloor \mathbf{v}^{\prime}_i
\end{bmatrix}
\begin{bmatrix}
\cos\theta_3\\
\sin\theta_3
\end{bmatrix}+({\mathbf{k}^{\prime}_3}^\sT\mathbf{v}^{\prime}_i)\mathbf{k}^{\prime}_3\right)=0\label{eq:At+b}
\end{align}
Expanding \eqref{eq:At+b} and rearranging terms, yields (for $ i=1,2 $)
\begin{align}
\label{eq:t1t3}
&\begin{bmatrix}
\cos\theta_1 \\ \sin\theta_1
\end{bmatrix}^\sT\begin{bmatrix}
\mathbf{u}_i^\sT\lfloor\mathbf{k}_1\rfloor^2\lfloor\mathbf{k}^{\prime}_3\rfloor^2\mathbf{v}^{\prime}_i & \mathbf{u}_i^\sT\lfloor\mathbf{k}_1\rfloor^2\lfloor\mathbf{k}^{\prime}_3\rfloor\mathbf{v}^{\prime}_i\\
\mathbf{u}_i^\sT\lfloor\mathbf{k}_1\rfloor\lfloor\mathbf{k}^{\prime}_3\rfloor^2\mathbf{v}^{\prime}_i &
\mathbf{u}_i^\sT\lfloor\mathbf{k}_1\rfloor\lfloor\mathbf{k}^{\prime}_3\rfloor\mathbf{v}^{\prime}_i
\end{bmatrix}\begin{bmatrix}
\cos\theta_3\\
\sin\theta_3
\end{bmatrix}\nonumber\\
+&(\mathbf{k}_1^\sT\mathbf{u}_i)\begin{bmatrix}
-\mathbf{k}_1^\sT\lfloor\mathbf{k}^{\prime}_3\rfloor^2\mathbf{v}^{\prime}_i & -\mathbf{k}_1^\sT\lfloor\mathbf{k}^{\prime}_3\rfloor\mathbf{v}^{\prime}_i
\end{bmatrix}\begin{bmatrix}
\cos\theta_3\\
\sin\theta_3
\end{bmatrix} \nonumber\\
=&({\mathbf{k}^{\prime}_3}^\sT\mathbf{v}^{\prime}_i)\begin{bmatrix}
\mathbf{u}_i^\sT\lfloor\mathbf{k}_1\rfloor\lfloor\mathbf{k}^{\prime}_3\rfloor\mathbf{k}_1 &
\mathbf{u}_i^\sT\lfloor\mathbf{k}_1\rfloor\mathbf{k}^{\prime}_3
\end{bmatrix}\begin{bmatrix}
\cos\theta_1\\
\sin\theta_1
\end{bmatrix}
\end{align}
Notice that the term $ \mathbf{u}_i^\sT\lfloor\mathbf{k}_1\rfloor\lfloor\mathbf{k}^{\prime}_3\rfloor$ appears three times in \eqref{eq:t1t3}, and
\begin{align}
\mathbf{u}_1^\sT\lfloor\mathbf{k}_1\rfloor\lfloor\mathbf{k}^{\prime}_3\rfloor&=\mathbf{u}_1^\sT\mathbf{k}^{\prime}_3\mathbf{k}_1^\sT\nonumber\\
&=({}^\sG\mathbf{p}_1-{}^\sG\mathbf{p}_3)^\sT\lfloor\mathbf{k}_1\rfloor\lfloor\mathbf{k}^{\prime}_3\rfloor\nonumber\\
&=({}^\sG\mathbf{p}_1-{}^\sG\mathbf{p}_2+{}^\sG\mathbf{p}_2-{}^\sG\mathbf{p}_3)^\sT\lfloor\mathbf{k}_1\rfloor\lfloor\mathbf{k}^{\prime}_3\rfloor\nonumber\\
&=({}^\sG\mathbf{p}_2-{}^\sG\mathbf{p}_3)^\sT\lfloor\mathbf{k}_1\rfloor\lfloor\mathbf{k}^{\prime}_3\rfloor\nonumber\\
&=\mathbf{u}_2^\sT\mathbf{k}^{\prime}_3\mathbf{k}_1^\sT= \mathbf{u}_2^\sT\lfloor\mathbf{k}_1\rfloor\lfloor\mathbf{k}^{\prime}_3\rfloor\label{eq:u1k3pk1}
\end{align}
This motivates to rewrite \eqref{eq:k3prime} as (for $ i=1,2 $):
\begin{align}
0&=\mathbf{u}_i^\sT\mathbf{C}(\mathbf{k}_1,\theta_1)\mathbf{C}(\mathbf{k}^\prime_3,\theta_3)\mathbf{v}^\prime_i\nonumber\\
&=\mathbf{u}_i^\sT\mathbf{C}(\mathbf{k}_1,\theta_1)\mathbf{C}(\mathbf{k}_1,-\phi)\mathbf{C}(\mathbf{k}_1,\phi)\mathbf{C}(\mathbf{k}^\prime_3,\theta_3)\mathbf{v}^\prime_i\nonumber\\
&=\mathbf{u}_i^\sT\mathbf{C}(\mathbf{k}_1,\theta_1-\phi)\mathbf{C}(\mathbf{C}(\mathbf{k}_1,\phi)\mathbf{k}^\prime_3,\theta_3)\mathbf{C}(\mathbf{k}_1,\phi)\mathbf{v}^\prime_i\nonumber\\
\label{eq:CuCv}&=\mathbf{u}_i^\sT\mathbf{C}(\mathbf{k}_1,\theta_1^\prime)\mathbf{C}(\mathbf{k}^{\prime\prime}_3,\theta_3)\mathbf{v}^{\prime\prime}_i
\end{align}
where
\begin{align}
\label{eq:t1p}\theta_1^\prime\triangleq\theta_1-\phi,\ \mathbf{v}^{\prime\prime}_i\triangleq\mathbf{C}(\mathbf{k}_1,\phi)\mathbf{v}^\prime_i,\ \mathbf{k}^{\prime\prime}_3\triangleq\mathbf{C}(\mathbf{k}_1,\phi)\mathbf{k}^\prime_3
\end{align}
To simplify the equation analogous to \eqref{eq:t1t3} that will result from \eqref{eq:CuCv} [instead of \eqref{eq:t1t3}], we seek to find a $ \phi $ (not necessarily unique) such that $ \mathbf{u}_1^\sT\mathbf{k}^{\prime\prime}_3=0 $, and hence, $ \mathbf{u}_i^\sT\lfloor\mathbf{k}_1\rfloor\lfloor\mathbf{k}^{\prime\prime}_3\rfloor=0 $ [see \eqref{eq:u1k3pk1}], \ie,
\begin{align}
0&=\mathbf{u}_1^\sT\mathbf{k}^{\prime\prime}_3=\mathbf{u}_1^\sT\mathbf{C}(\mathbf{k}_1,\phi)\mathbf{k}^\prime_3\label{eq:u1k3pp}\\
&=\mathbf{u}_1^\sT(\cos\phi\mathbf{I}-\sin\phi\lfloor\mathbf{k}_1\rfloor+(1-\cos\phi)\mathbf{k}_1\mathbf{k}_1^\sT)\mathbf{k}^\prime_3\nonumber\\
&=\cos\phi\mathbf{u}_1^\sT\mathbf{k}^\prime_3-\sin\phi\mathbf{u}_1^\sT\lfloor\mathbf{k}_1\rfloor\mathbf{k}^\prime_3\nonumber\\
&=\cos\phi\mathbf{u}_1^\sT\mathbf{k}^\prime_3-\sin\phi\mathbf{u}_1^\sT\mathbf{k}_2\nonumber
\end{align}
\begin{align}
\label{eq:phi}\Rightarrow\begin{bmatrix}
\cos\phi & \sin\phi
\end{bmatrix}&=\frac{1}{\delta}\begin{bmatrix}
\mathbf{u}_1^\sT\mathbf{k}_2 & \mathbf{u}_1^\sT\mathbf{k}^\prime_3
\end{bmatrix}
\end{align}
where
\begin{align}
\label{eq:delta}\delta\triangleq\sqrt{(\mathbf{u}_1^\sT\mathbf{k}^\prime_3)^2+(\mathbf{u}_1^\sT\mathbf{k}_2)^2}
=\|\mathbf{u}_1\times\mathbf{k}_1\|
\end{align}
and thus [from \eqref{eq:t1p} using \eqref{eq:Rod}]
\begin{align}
\mathbf{k}_3^{\prime\prime}&=\cos\phi\mathbf{k}^\prime_3-\sin\phi\lfloor\mathbf{k}_1\rfloor\mathbf{k}^\prime_3+(1-\cos\phi)\mathbf{k}_1\mathbf{k}_1^\sT\mathbf{k}^\prime_3\nonumber\\
&=(\mathbf{k}^\prime_3\mathbf{k}_2^\sT\mathbf{u}_1-\mathbf{k}_2{\mathbf{k}_3^\prime}^\sT\mathbf{u}_1)/\delta=\mathbf{u}_1\times(\mathbf{k}^\prime_3\times\mathbf{k}_2)/\delta\nonumber\\
\label{eq:k3pp}&=\frac{\mathbf{u}_1\times\mathbf{k}_1}{\|\mathbf{u}_1\times\mathbf{k}_1\|}
\end{align}
Now, we can expand \eqref{eq:CuCv} using \eqref{eq:Cv} to get an equation analogous to \eqref{eq:t1t3}:
\begin{align}
&\begin{bmatrix}
\cos\theta_1^\prime \\ \sin\theta_1^\prime
\end{bmatrix}^\sT\begin{bmatrix}
\mathbf{u}_i^\sT\lfloor\mathbf{k}_1\rfloor^2\lfloor\mathbf{k}^{\prime\prime}_3\rfloor^2\mathbf{v}^{\prime\prime}_i & \mathbf{u}_i^\sT\lfloor\mathbf{k}_1\rfloor^2\lfloor\mathbf{k}^{\prime\prime}_3\rfloor\mathbf{v}^{\prime\prime}_i \nonumber\\
\mathbf{u}_i^\sT\lfloor\mathbf{k}_1\rfloor\lfloor\mathbf{k}^{\prime\prime}_3\rfloor^2\mathbf{v}^{\prime\prime}_i &
\mathbf{u}_i^\sT\lfloor\mathbf{k}_1\rfloor\lfloor\mathbf{k}^{\prime\prime}_3\rfloor\mathbf{v}^{\prime\prime}_i
\end{bmatrix}\begin{bmatrix}
\cos\theta_3\\
\sin\theta_3
\end{bmatrix}\nonumber\\
&+(\mathbf{k}_1^\sT\mathbf{u}_i)\begin{bmatrix}
-\mathbf{k}_1^\sT\lfloor\mathbf{k}^{\prime\prime}_3\rfloor^2\mathbf{v}^{\prime\prime}_i & -\mathbf{k}_1^\sT\lfloor\mathbf{k}^{\prime\prime}_3\rfloor\mathbf{v}^{\prime\prime}_i
\end{bmatrix}\begin{bmatrix}
\cos\theta_3\\
\sin\theta_3
\end{bmatrix}=\nonumber\\
\label{eq:matrix}&({\mathbf{k}^{\prime\prime}_3}^\sT\mathbf{v}^{\prime\prime}_i)\begin{bmatrix}
\mathbf{u}_i^\sT\lfloor\mathbf{k}_1\rfloor\lfloor\mathbf{k}^{\prime\prime}_3\rfloor\mathbf{k}_1 &
\mathbf{u}_i^\sT\lfloor\mathbf{k}_1\rfloor\mathbf{k}^{\prime\prime}_3
\end{bmatrix}\begin{bmatrix}
\cos\theta_1^\prime\\
\sin\theta_1^\prime
\end{bmatrix}
\end{align}
Substituting $ \mathbf{u}_i^\sT\lfloor\mathbf{k}_1\rfloor\lfloor\mathbf{k}^{\prime\prime}_3\rfloor=0 $ [see \eqref{eq:u1k3pk1}] in \eqref{eq:matrix} and renaming terms, yields (for $ i=1,2 $):
\begin{align}
&\begin{bmatrix}
\cos\theta_1^\prime \\ \sin\theta_1^\prime
\end{bmatrix}^\sT\begin{bmatrix}
\bar{f}_{i1} & \bar{f}_{i2}\\
0 & 0
\end{bmatrix}\begin{bmatrix}
\cos\theta_3\\
\sin\theta_3
\end{bmatrix}+\begin{bmatrix}
\bar{f}_{i4} & \bar{f}_{i5}
\end{bmatrix}\begin{bmatrix}
\cos\theta_3\\
\sin\theta_3
\end{bmatrix}\nonumber\\
=&\begin{bmatrix}
0 & \bar{f}_{i3}
\end{bmatrix}\begin{bmatrix}
\cos\theta_1^\prime\\
\sin\theta_1^\prime
\end{bmatrix}\nonumber\\
\label{eq:fi}
\Rightarrow
&\begin{bmatrix}
\bar{f}_{i1}\cos\theta_1^\prime+\bar{f}_{i4} & \bar{f}_{i2}\cos\theta_1^\prime+\bar{f}_{i5}
\end{bmatrix}\begin{bmatrix}
\cos\theta_3\\
\sin\theta_3
\end{bmatrix}=\bar{f}_{i3}\sin\theta_1^\prime
\end{align}
where\footnote{The simplified expressions for the following terms, shown after the second equality, require lengthy algebraic derivations which we omit due to space limitations.}
\begin{align}
\bar{f}_{i1}&\triangleq\mathbf{u}_i^\sT\lfloor\mathbf{k}_1\rfloor^2\lfloor\mathbf{k}^{\prime\prime}_3\rfloor^2\mathbf{v}^{\prime\prime}_i=\delta\mathbf{v}_i^\sT\mathbf{k}_2\nonumber\\
\bar{f}_{i2}&\triangleq\mathbf{u}_i^\sT\lfloor\mathbf{k}_1\rfloor^2\lfloor\mathbf{k}^{\prime\prime}_3\rfloor\mathbf{v}^{\prime\prime}_i=\delta\mathbf{v}_i^\sT\mathbf{k}_1^\prime\nonumber\\
\bar{f}_{i3}&\triangleq({\mathbf{k}^{\prime\prime}_3}^\sT\mathbf{v}^{\prime\prime}_i)\mathbf{u}_i^\sT\lfloor\mathbf{k}_1\rfloor\mathbf{k}^{\prime\prime}_3=\delta\mathbf{v}_i^\sT\mathbf{k}_3\nonumber\\
\bar{f}_{i4}&\triangleq-(\mathbf{u}_i^\sT\mathbf{k}_1)\mathbf{k}_1^\sT\lfloor\mathbf{k}^{\prime\prime}_3\rfloor^2\mathbf{v}^{\prime\prime}_i=(\mathbf{u}_i^\sT\mathbf{k}_1)(\mathbf{v}_i^\sT\mathbf{k}_1^\prime)\nonumber\\
\bar{f}_{i5}&\triangleq-(\mathbf{u}_i^\sT\mathbf{k}_1)\mathbf{k}_1^\sT\lfloor\mathbf{k}^{\prime\prime}_3\rfloor\mathbf{v}^{\prime\prime}_i=-(\mathbf{u}_i^\sT\mathbf{k}_1)(\mathbf{v}_i^\sT\mathbf{k}_2)\nonumber
\end{align}
For $ i=1,2 $, \eqref{eq:fi} results into the following system:
\begin{equation}
\begin{bmatrix}
\label{eq:Fi}
\bar{f}_{11}\cos\theta_1^\prime+\bar{f}_{14} & \bar{f}_{12}\cos\theta_1^\prime+\bar{f}_{15} \\
\bar{f}_{21}\cos\theta_1^\prime+\bar{f}_{24} & \bar{f}_{22}\cos\theta_1^\prime+\bar{f}_{25}
\end{bmatrix}\begin{bmatrix}
\cos\theta_3\\
\sin\theta_3
\end{bmatrix}=\begin{bmatrix}
\bar{f}_{13}\\
\bar{f}_{23}
\end{bmatrix}\sin\theta_1^\prime
\end{equation}
Note that since $ \bar{f}_{11}\bar{f}_{14}+\bar{f}_{12}\bar{f}_{15}=0 $, we can further simplify \eqref{eq:Fi} by introducing $ \theta_3^\prime $, where
\begin{align}
\label{eq:t3p}\begin{bmatrix}
\cos\theta_3^\prime\\
\sin\theta_3^\prime
\end{bmatrix}\triangleq\begin{bmatrix}\frac{\bar{f}_{11}\cos\theta_3+\bar{f}_{12}\sin\theta_3}{\sqrt{\bar{f}_{11}^2+\bar{f}_{12}^2}} & -\frac{\bar{f}_{14}\cos\theta_3+\bar{f}_{15}\sin\theta_3}{\sqrt{\bar{f}_{14}^2+\bar{f}_{15}^2}}\end{bmatrix}^\sT
\end{align}
Replacing $ \theta_3 $ by $ \theta_3^\prime $ in \eqref{eq:Fi}, we have
\begin{equation}
\begin{bmatrix}
f_{11}\cos\theta_1^\prime & f_{15} \\
f_{21}\cos\theta_1^\prime+f_{24} & f_{22}\cos\theta_1^\prime+f_{25}
\end{bmatrix}\begin{bmatrix}
\cos\theta_3^\prime\\
\sin\theta_3^\prime
\end{bmatrix}=\begin{bmatrix}
f_{13}\\
f_{23}
\end{bmatrix}\sin\theta_1^\prime\label{eq:fs1}
\end{equation}
where
\begin{align}
\label{eq:fi1}f_{11}&\triangleq\delta\mathbf{k}_3^\sT{}^\sCn\mathbf{b}_3\\
f_{21}&\triangleq\delta({}^\sCn\mathbf{b}_1^\sT{}^\sCn\mathbf{b}_2)(\mathbf{k}_3^\sT{}^\sCn\mathbf{b}_3)\\
f_{22}&\triangleq\delta(\mathbf{k}_3^\sT{}^\sCn\mathbf{b}_3)\|{}^\sCn\mathbf{b}_1\times{}^\sCn\mathbf{b}_2\|\\
f_{13}&\triangleq\bar{f}_{13}=\delta\mathbf{v}_1^\sT\mathbf{k}_3\\
f_{23}&\triangleq\bar{f}_{23}=\delta\mathbf{v}_2^\sT\mathbf{k}_3\\
f_{24}&\triangleq(\mathbf{u}_2^\sT\mathbf{k}_1)(\mathbf{k}_3^\sT{}^\sCn\mathbf{b}_3)\|{}^\sCn\mathbf{b}_1\times{}^\sCn\mathbf{b}_2\|\\
f_{15}&\triangleq-(\mathbf{u}_1^\sT\mathbf{k}_1)(\mathbf{k}_3^\sT{}^\sCn\mathbf{b}_3)\\
\label{eq:fi5}f_{25}&\triangleq-(\mathbf{u}_2^\sT\mathbf{k}_1)({}^\sCn\mathbf{b}_1^\sT{}^\sCn\mathbf{b}_2)(\mathbf{k}_3^\sT{}^\sCn\mathbf{b}_3)
\end{align}
From \eqref{eq:fs1}, we have
\begin{align}
\label{eq:theta3p}
\begin{bmatrix}
\cos\theta_3^\prime\\
\sin\theta_3^\prime
\end{bmatrix}=&\det\left(\begin{bmatrix}
f_{11}\cos\theta_1^\prime & f_{15} \\
f_{21}\cos\theta_1^\prime+f_{24} & f_{22}\cos\theta_1^\prime+f_{25}
\end{bmatrix}\right)^{-1}\nonumber\\
\cdot&\begin{bmatrix}
f_{22}\cos\theta_1^\prime+f_{25} & -f_{15} \\
-(f_{21}\cos\theta_1^\prime+f_{24}) & f_{11}\cos\theta_1^\prime
\end{bmatrix}\begin{bmatrix}
f_{13}\\
f_{23}
\end{bmatrix}\sin\theta_1^\prime
\end{align}
Computing the norm of both sides of \eqref{eq:theta3p}, results in
\begin{align}
&\left\lVert\begin{bmatrix}
f_{22}\cos\theta_1^\prime+f_{25} & -f_{15} \\
-(f_{21}\cos\theta_1^\prime+f_{24}) & f_{11}\cos\theta_1^\prime
\end{bmatrix}\begin{bmatrix}
f_{13}\\
f_{23}
\end{bmatrix}\right\rVert^2(1-\cos^2\theta_1^\prime)\nonumber\\
&=\det\left(\begin{bmatrix}
f_{11}\cos\theta_1^\prime & f_{15} \\
f_{21}\cos\theta_1^\prime+f_{24} & f_{22}\cos\theta_1^\prime+f_{25}
\end{bmatrix}\right)^2\nonumber
\end{align}
which is a 4th-order polynomial in $ \cos\theta_1^\prime $ that can be compactly written as:
\begin{equation}
\label{eq:4th}\displaystyle\sum_{j=0}^{4}\alpha_j\cos^j\theta_1^\prime=0
\end{equation}
with
\begin{align}
\label{eq:A4}\alpha_4&\triangleq g_5^2+g_1^2+g_3^2\\
\alpha_3&\triangleq 2(g_5g_6+g_1g_2+g_3g_4)\\
\alpha_2&\triangleq g_6^2+2g_5g_7+g_2^2+g_4^2-g_1^2-g_3^2\\
\alpha_1&\triangleq 2(g_6g_7-g_1g_2-g_3g_4)\\
\alpha_0&\triangleq g_7^2-g_2^2-g_4^2\\
g_1&\triangleq f_{13}f_{22}\\
g_2&\triangleq f_{13}f_{25}-f_{15}f_{23}\\
g_3&\triangleq f_{11}f_{23}-f_{13}f_{21}\\
g_4&\triangleq -f_{13}f_{24}\\
g_5&\triangleq f_{11}f_{22}\\
g_6&\triangleq f_{11}f_{25}-f_{15}f_{21}\\
\label{eq:g7}g_7&\triangleq -f_{15}f_{24}
\end{align}
We compute the roots of \eqref{eq:4th} in closed form to find $ \cos\theta_1^\prime $. Similarly to~\cite{kneip2011novel} and~\cite{masselli2014new}, we employ Ferrari's method~\cite{cardano2007rules} to attain the resolvent cubic of \eqref{eq:4th}, which is subsequently solved by Cardano's formula~\cite{cardano2007rules}. 
Once the (up to) four real solutions of \eqref{eq:4th} have been determined, an optional step is to apply root polishing following Newton's method, which improves accuracy for minimal increase in the processing cost (see Section~\ref{ssec:cost}). 
Regardless, for each solution of $ \cos\theta_1^\prime $, we will have two possible solutions for $ \sin\theta_1^\prime $, \ie,
\begin{equation}
\label{eq:st1p}\sin\theta_1^\prime=\pm\sqrt{1-\cos^2\theta_1^\prime}
\end{equation}
which, in general, will result in two different solutions for $ {}^\sCn_\sG\mathbf{C} $. Note though that only one of them is valid if we use the fact that $ d_i>0 $ (see Appendix~\ref{ssec:stheta1p}). 

\indent Next, for each pair of $ (\cos\theta_1^\prime,\sin\theta_1^\prime)$, we compute $ \cos\theta_3^\prime$ and $\sin\theta_3^\prime $ from \eqref{eq:theta3p}, which can be written as
\begin{equation}
\label{eq:t3}
\begin{bmatrix}
\cos \theta_3^\prime\\
\sin \theta_3^\prime
\end{bmatrix}=\frac{\sin\theta_1^\prime}{g_5\cos^2\theta_1^\prime+g_6\cos\theta_1^\prime+g_7}
\begin{bmatrix}
g_1\cos\theta_1^\prime+g_2\\
g_3\cos\theta_1^\prime+g_4
\end{bmatrix}
\end{equation}
\indent Lastly, instead of first computing $ \theta_1 $ from \eqref{eq:t1p} and $ \theta_3 $ from \eqref{eq:t3p} to find $ {}^\sG_\sCn\mathbf{C} $ using \eqref{eq:ccc}, we hereafter describe a faster method for recovering $ {}^\sG_\sCn\mathbf{C} $. Specifically, from \eqref{eq:ccc}, \eqref{eq:k3prime} and \eqref{eq:CuCv}, we have
\begin{align}
{}^\sG_\sCn\mathbf{C}&=\mathbf{C}(\mathbf{k}_1,\theta_1)\mathbf{C}(\mathbf{k}_2,\theta_2)\mathbf{C}(\mathbf{k}_3,\theta_3)\nonumber\\
&=\mathbf{C}(\mathbf{k}_1,\theta_1)\mathbf{C}(\mathbf{k}_3^\prime,\theta_3)\mathbf{C}(\mathbf{k}_2,\theta_2)\nonumber\\
&=\mathbf{C}(\mathbf{k}_1,\theta_1^\prime)\mathbf{C}(\mathbf{k}_3^{\prime\prime},\theta_3)\mathbf{C}(\mathbf{k}_1,\phi)\mathbf{C}(\mathbf{k}_2,\theta_2)\label{eq:cold}
\end{align}
Since $ \mathbf{k}_1 $ is perpendicular to $ \mathbf{k}_3^{\prime\prime} $, we can construct a rotation matrix $ \mathbf{\bar{C}} $ such that
\begin{align}
\mathbf{\bar{C}}=\begin{bmatrix}
\mathbf{k}_1 & \mathbf{k}_3^{\prime\prime} & \mathbf{k}_1\times\mathbf{k}_3^{\prime\prime}
\end{bmatrix}\nonumber
\end{align}
and hence
\begin{align}
\label{eq:kcbar}\mathbf{k}_1=\mathbf{\bar{C}}\mathbf{e}_1,\ \mathbf{k}_3^{\prime\prime}=\mathbf{\bar{C}}\mathbf{e}_2
\end{align}
where
\begin{align}
\begin{bmatrix}
\mathbf{e}_1 & \mathbf{e}_2 & \mathbf{e}_3
\end{bmatrix}\triangleq\mathbf{I}_3\nonumber
\end{align}
Substituting \eqref{eq:kcbar} in \eqref{eq:cold}, we have
\begin{align}
{}^\sG_\sCn\mathbf{C}&=\mathbf{\bar{C}}\mathbf{C}(\mathbf{e}_1,\theta_1^\prime)\mathbf{C}(\mathbf{e}_2,\theta_3)\mathbf{\bar{C}}^\sT\mathbf{C}(\mathbf{k}_1,\phi)\mathbf{C}(\mathbf{k}_2,\theta_2)\nonumber\\
&=\mathbf{\bar{C}}\mathbf{C}(\mathbf{e}_1,\theta_1^\prime)\mathbf{C}(\mathbf{e}_2,\theta_3)\mathbf{C}(\mathbf{e}_2,\theta_3^\prime-\theta_3)\mathbf{\bar{\bar{C}}}\nonumber\\
\label{eq:crrc}&=\mathbf{\bar{C}}\mathbf{C}(\mathbf{e}_1,\theta_1^\prime)\mathbf{C}(\mathbf{e}_2,\theta_3^\prime)\mathbf{\bar{\bar{C}}}
\end{align}
where
\begin{align}
\mathbf{\bar{\bar{C}}}&\triangleq\mathbf{C}(\mathbf{e}_2,\theta_3-\theta_3^\prime)\mathbf{\bar{C}}^\sT\mathbf{C}(\mathbf{k}_1,\phi)\mathbf{C}(\mathbf{k}_2,\theta_2)\nonumber\\
&=\mathbf{C}(\mathbf{e}_2,\theta_3-\theta_3^\prime)\begin{bmatrix}
\mathbf{k}_1^\prime & \mathbf{k}_3 & \mathbf{k}_1^\prime\times\mathbf{k}_3
\end{bmatrix}^\sT\nonumber\\
&\stackrel{\eqref{eq:t3p}}{=}\begin{bmatrix}
{}^\sCn\mathbf{b}_1 & \mathbf{k}_3 & {}^\sCn\mathbf{b}_1\times\mathbf{k}_3
\end{bmatrix}^\sT\nonumber
\end{align}
The advantages of \eqref{eq:crrc} are: (i) The matrix product $ \mathbf{C}(\mathbf{e}_1,\theta_1^\prime)\mathbf{C}(\mathbf{e}_2,\theta_3^\prime) $ can be computed analytically; (ii) $ \mathbf{\bar{C}},\mathbf{\bar{\bar{C}}} $ are invariant to the (up to) four possible solutions and thus, we only need to construct them once. 
\subsection{Solving for the position}
Substituting in \eqref{eq:di} the expression for $ d_3 $ from \eqref{eq:s1pd3} and rearranging terms, yields
\begin{align}
\label{eq:gpc} {}^\sG\mathbf{p}_\sCn&={}^\sG\mathbf{p}_3-\frac{\delta\sin\theta_1^\prime}{\mathbf{k}_3^\sT{}^\sCn\mathbf{b}_3}{}^\sG_\sCn\mathbf{C}{}^\sCn\mathbf{b}_3
\end{align}
Note that we only use \eqref{eq:di} for $ i=3 $ to compute $ {}^\sG\mathbf{p}_\sCn $ from $ {}^\sG_\sCn\mathbf{C} $. 
Alternatively, we could find the position using a least-squares approach based on \eqref{eq:di} for $ i=1,2,3 $ (see Appendix~\ref{ssec:ls}), if we care more for accuracy than speed. 
Lastly, the proposed P3P solution is summarized in Alg.~\ref{alg:p3p}.
\begin{Ualgorithm}
\KwIn{$ {}^\sG\mathbf{p}_i,~i = 1,2,3 $ the features' positions; $ {}^\sCn\mathbf{b}_i,~i = 1,2,3 $ bearing measurements}
\KwOut{$ {}^\sG\mathbf{p}_\sCn $, the position of the camera; $ {}^\sCn_\sG\mathbf{C} $, the orientation of the camera}
Compute $ \mathbf{k}_1$, $ \mathbf{k}_3$ using \eqref{eq:ki}\\
Compute $ \mathbf{u}_i $ and $ \mathbf{v}_i $ using \eqref{eq:ui}, $ i=1,2 $\\
Compute $ \delta $ and $ \mathbf{k}_3^{\prime\prime} $ using \eqref{eq:delta} and \eqref{eq:k3pp}\\
Compute the $ f_{ij} $'s using \eqref{eq:fi1}-\eqref{eq:fi5}\\
Compute $ \alpha_i $, $ i=0,1,2,3,4 $ using \eqref{eq:A4}-\eqref{eq:g7}\\
Solve \eqref{eq:4th} to get $ n$ ($ n=2 $ or $ 4 $) real solutions for $ \cos\theta_1^\prime $, denoted as $ \cos\theta_1^{\prime(i)} $, $ i=1...n $\\
\For {$i = 1:n$}{
%\eIf {$ \mathbf{k}_3^\sT{}^\sCn\mathbf{b}_3>0 $}{
%$ \sin\theta_1^{\prime(i)}\gets\sqrt{1-\cos^2\theta_1^{\prime(i)}} $}
%{
%$ \sin\theta_1^{\prime(i)}\gets-\sqrt{1-\cos^2\theta_1^{\prime(i)}} $}
$ \sin\theta_1^{\prime(i)}\gets \sign(\mathbf{k}_3^\sT{}^\sCn\mathbf{b}_3) \sqrt{1-\cos^2\theta_1^{\prime(i)}} $\\
Compute $ \cos\theta_3^{\prime(i)} $ and $ \sin\theta_3^{\prime(i)} $ using \eqref{eq:t3}\\
Compute $ {}^\sCn_\sG\mathbf{C}^{(i)} $ using \eqref{eq:crrc}\\
Compute $ {}^\sG\mathbf{p}_\sCn^{(i)} $ using \eqref{eq:gpc}\\
}
\caption{Solving for the camera's pose}
\label{alg:p3p}
\end{Ualgorithm}
%%%%%%%%%%%%%%%%%%%%%%%%%%%%%%%%%%%%%%%%%%%%%%%%%%%%
\section{Experimental results}
\label{sec:results}
%%%%%%%%%%%%%%%%%%%%%%%%%%%%%%%%%%%%%%%%%%%%%%%%%%%%
Our algorithm is implemented\footnote{Our code is submitted along with the paper as supplemental material.} in C++ using the same linear algebra library, TooN~\cite{TooN_library}, as~\cite{kneip2011novel}.
We employ simulation data to test our code and compare it to the solutions of~\cite{kneip2011novel}~and~\cite{masselli2014new}.
For each single P3P problem, we randomly generate three 3D landmarks, which are uniformly distributed in a $ 0.4\times0.3\times0.4 $ cuboid centered around the origin. 
The position of the camera is $ {}^\sG\mathbf{p}_\sCn=\mathbf{e}_3 $, and its orientation is $ {}^\sCn_\sG\mathbf{C}=\mathbf{C}(\mathbf{e}_1,\pi) $.
\subsection{Numerical accuracy} 
\label{ssec:numerical}
We generate simulation data without adding any noise or rounding error to the bearing measurements, and run all three algorithms on 50,000 randomly-generated configurations to assess their numerical accuracy. 
Note that the position error is computed as the norm of the difference between the estimate and the ground truth of $ {}^\sG\mathbf{p}_\sCn $. 
As for the orientation error, we compute the rotation matrix that transforms the estimated $ {}^\sG_\sCn\mathbf{C} $ to the true one, convert it to the equivalent axis-angle representation, and use the absolute value of the angle as the error. 
Since there are multiple solutions to a P3P problem, we compute the errors for all of them and pick the smallest one (\ie, the root closest to the true solution).

\begin{figure}
\center
\includegraphics[width=\linewidth]{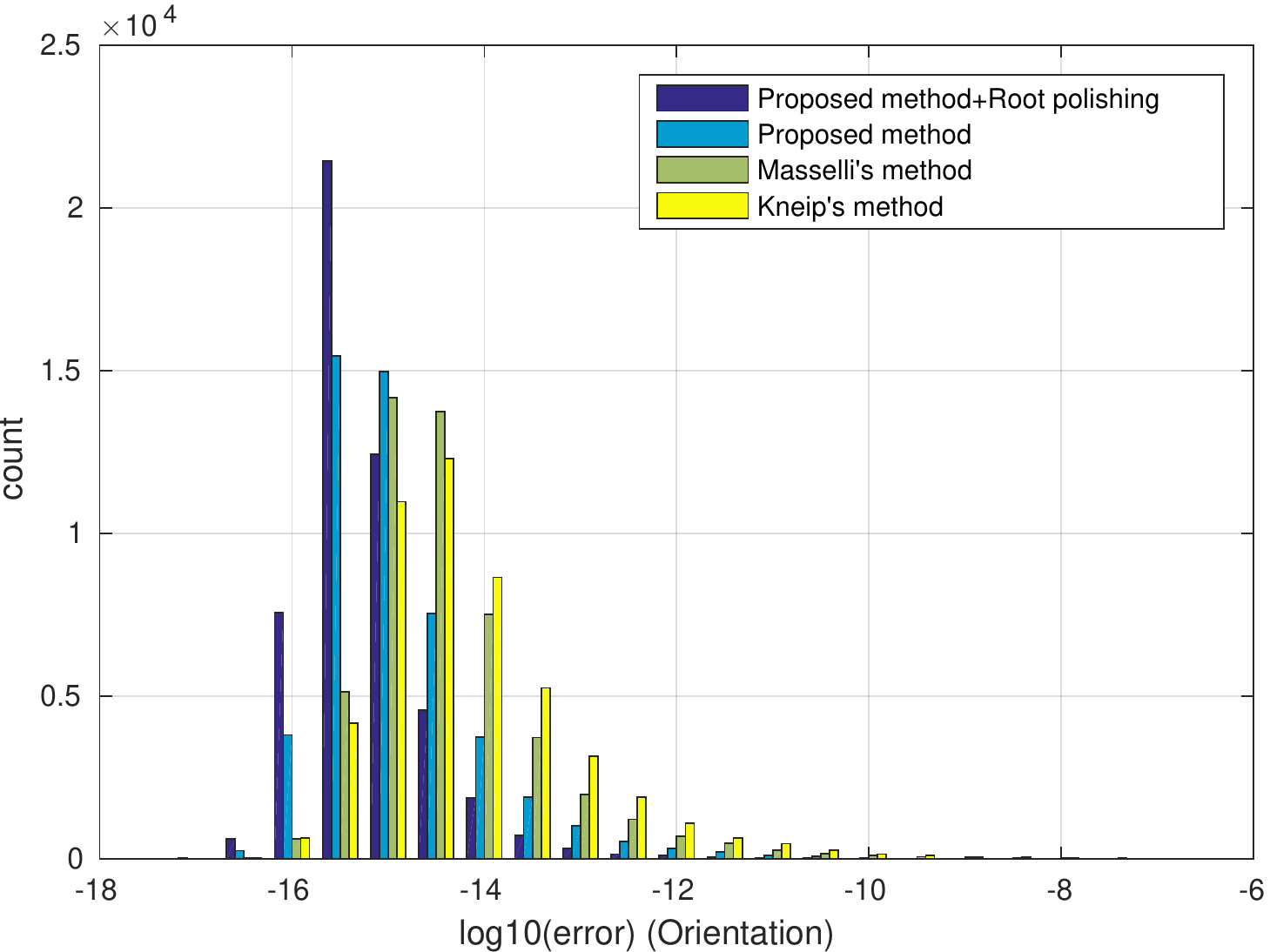}
\caption{Nominal case: Histogram of orientation errors.}
\label{fig:errorO}
\end{figure}

\begin{figure}
\center
\includegraphics[width=\linewidth]{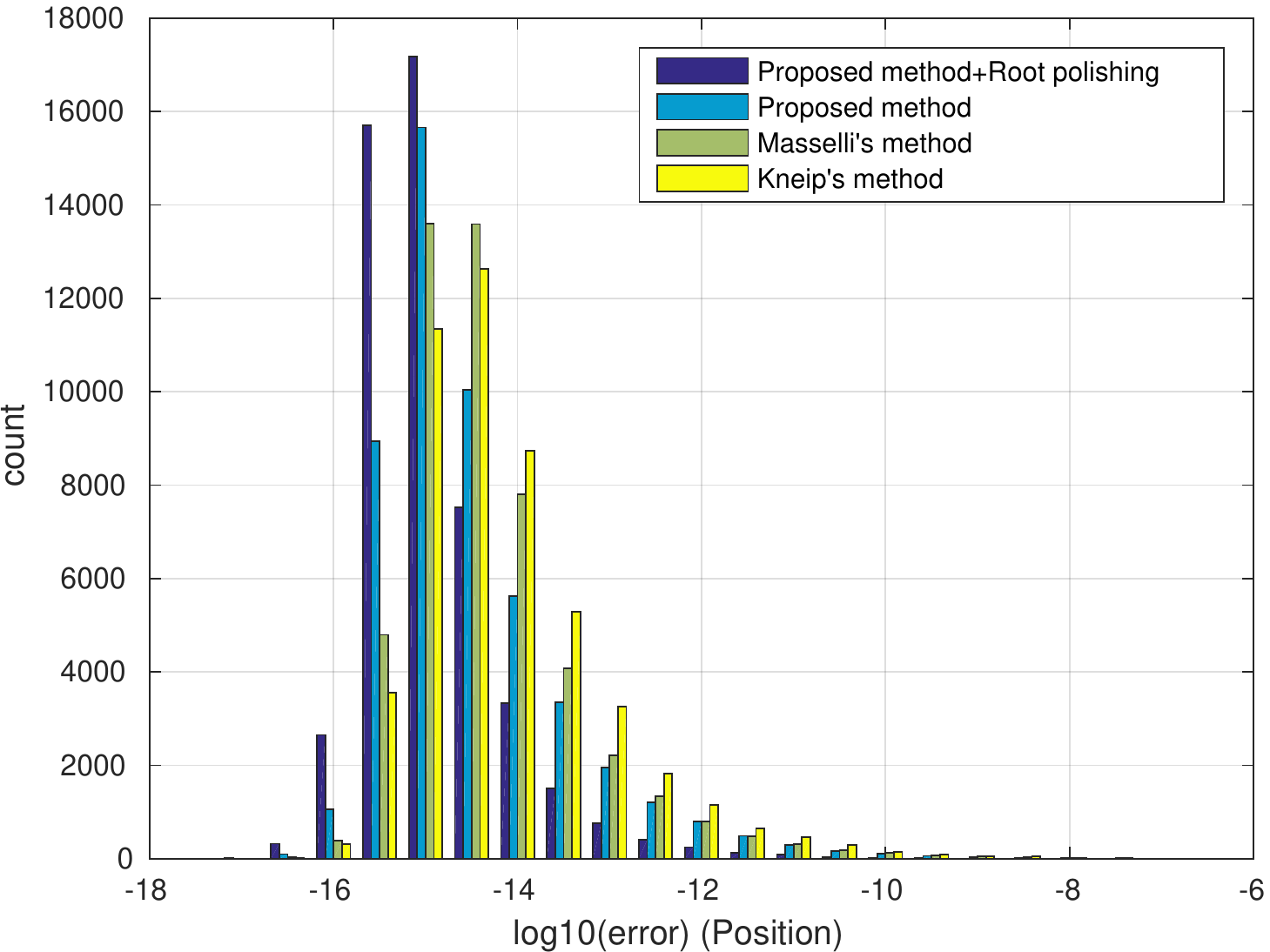}
\caption{Nominal case: Histogram of position errors.}
\label{fig:errorP}
\end{figure}

The distributions and the means of the position and orientation errors are depicted in Fig.s~\ref{fig:errorO}~-~\ref{fig:errorP} and Table~\ref{tab:err}. 
As evident, we get similar results to those presented in~\cite{masselli2014new} for Kneip~\etal's~\cite{kneip2011novel} and Masselli and Zell's methods~\cite{masselli2014new}, while our approach outperforms both of them by two orders of magnitude in terms of accuracy.
This can be attributed to the fact that our algorithm requires fewer operations and thus exhibits lower numerical-error propagation.

Furthermore, and as shown in the results of Table~\ref{tab:err}, we can further improve the numerical precision by applying root polishing. 
Typically, two iterations of Newton's method~\cite{ypma1995historical}  lead to significantly better results, especially for the orientation, while taking only 0.01 $\mu$s per iteration, or about 4\% of the total processing time.
%

%
%Furthermore, and as shown in the results of Table~\ref{tab:err}, we can further improve the numerical precision by applying root polishing. 
%%
%According to our speed tests, root polishing costs only about 0.01 $\mu$s per iteration. 
%%
%Typically, two iterations of Newton's method~\cite{ypma1995historical}  lead to significantly better results, especially for the orientation, while taking only about 4\% of the total processing time.
%%

\begin{table}
\label{tab:err}
\center
\begin{tabular}{|c|c|c|}
\hline
 & position & orientation  \\
\hline
Kneip's method & 1.18E-05 & 1.02E-05 \\
\hline
Masselli's method & 1.84E-08 & 4.89E-10\\
\hline
Proposed method & \bf1.66E-10 & \bf5.30E-12 \\
\hline
Proposed method+Root polishing & \bf5.07E-11 & \bf1.53E-13 \\
\hline
\end{tabular}
  \caption{Nominal case: Pose mean errors.}
\end{table}
\subsection{Processing cost} 
\label{ssec:cost}
We use a test program that solves 100,000 randomly generated P3P problems and calculates the total execution time to evaluate the computational cost of the three algorithms considered. 
We run it on a 2.0 GHz$ \times $4 Core laptop and the results show that our code takes 0.54 $ \mu $s on average (0.52 $ \mu $s without root polishing) while~\cite{kneip2011novel} and~\cite{masselli2014new} take 1.3 $ \mu $s and 1.5 $ \mu $s, respectively. 
This corresponds to a 2.5$\times$ speed up (or 40\% of the time of~\cite{kneip2011novel}).
Note also, in contrast to what is reported in~\cite{masselli2014new}, Masselli's method is actually slower than Kneip's. 
As mentioned earlier, Masselli's results in~\cite{masselli2014new} are based on 1,000 runs of the same features' configuration, and take advantage of data caching to outperform Kneip.
%; this was not the case for the tests us Kneip's code.

%
%LALA
%
%Our code takes approximately 40\% of the time of~\cite{kneip2011novel}. Notice that Masselli's code is actually slower than Kneip's. 
%%
%As mentioned before, Masselli's experiments in~\cite{masselli2014new} are based on 1,000 runs of the same setting of 3 landmarks, so they are able to outperform Kneip because of caching in their data only.
%

\subsection{Robustness}
\begin{figure}
\center
\includegraphics[width=\linewidth]{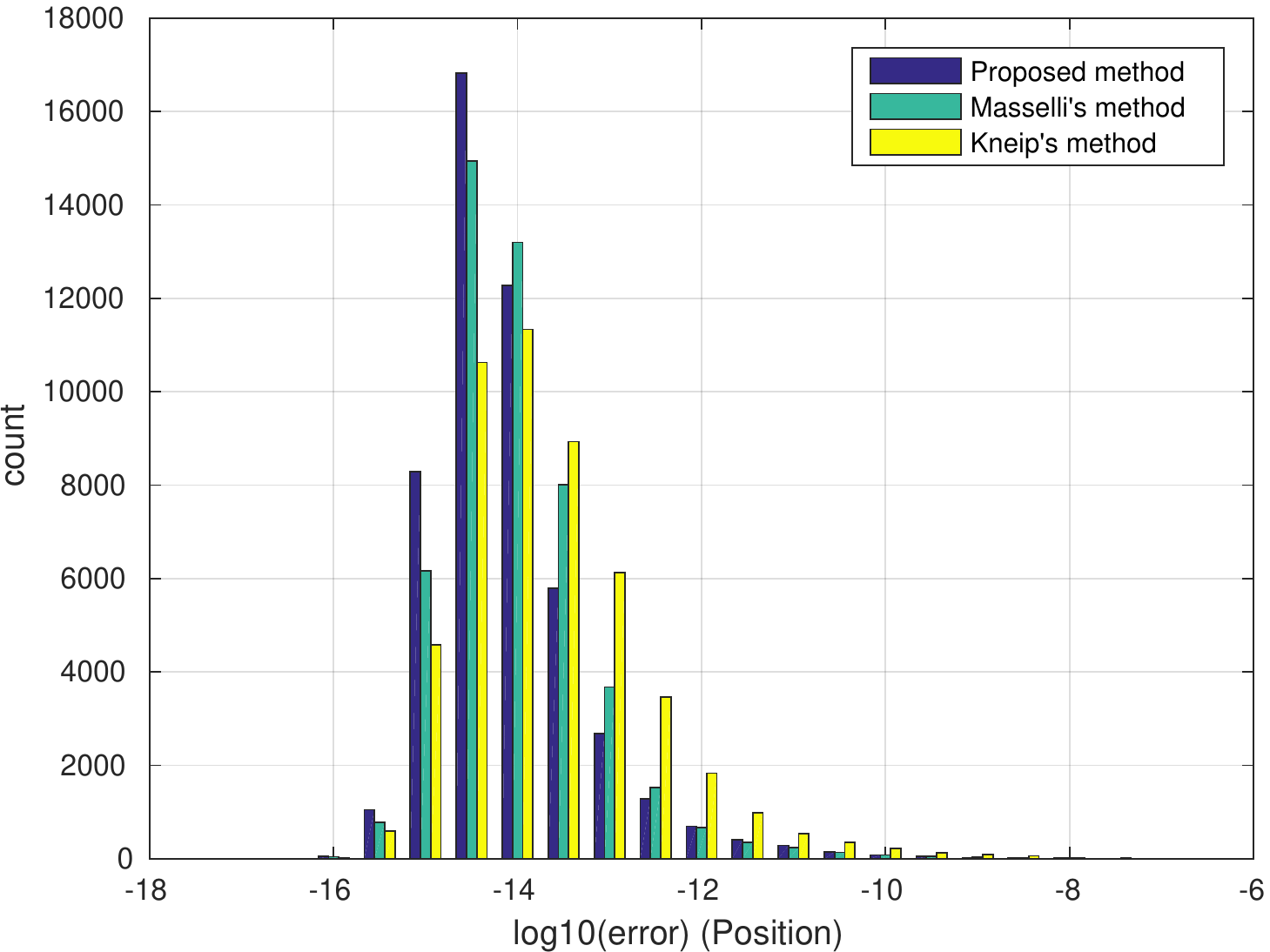}
\caption{Singular case 1: Histogram of position errors.}
\label{fig:robustp1}
\end{figure}

\begin{figure}
\center
\includegraphics[width=\linewidth]{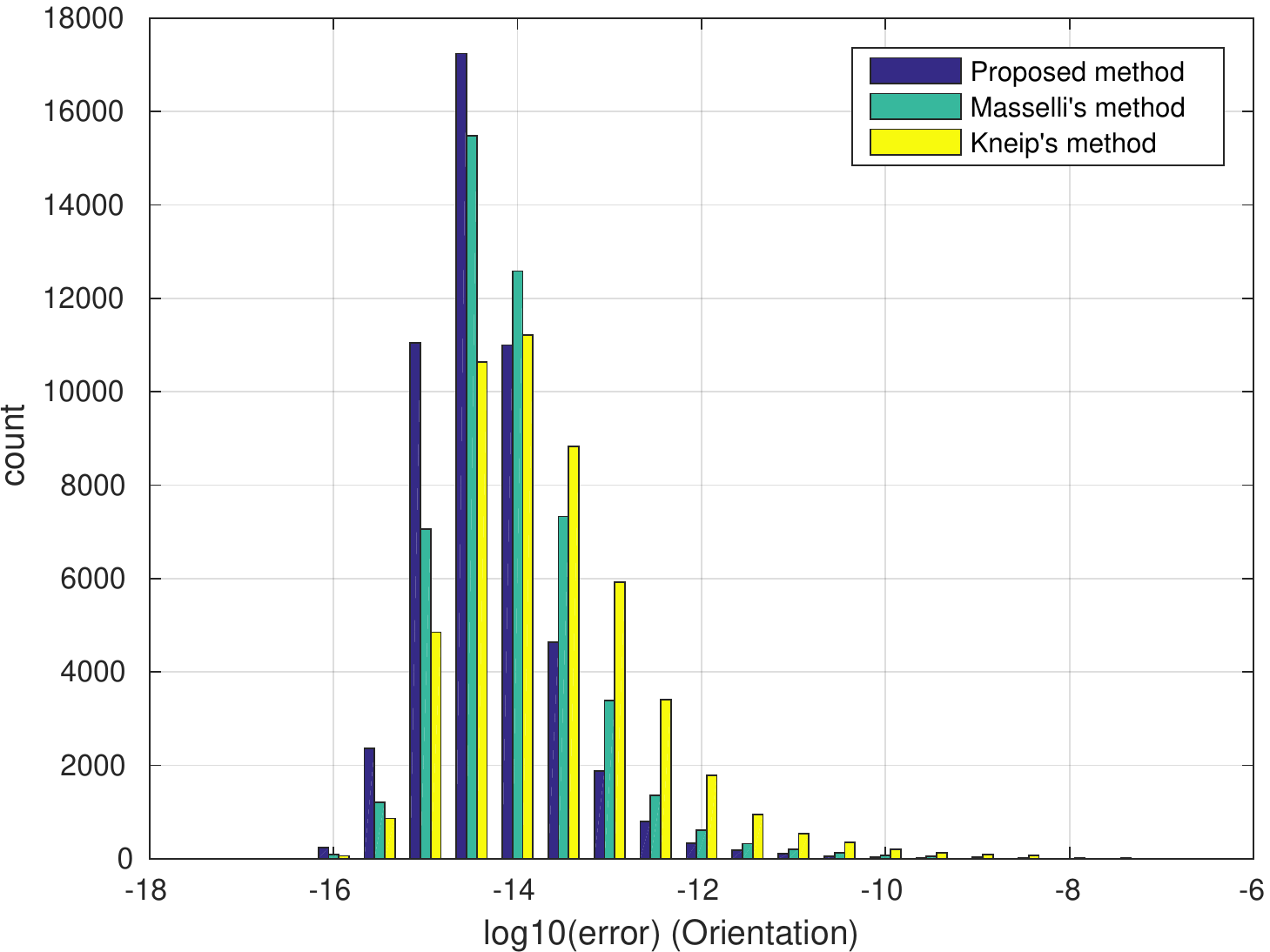}
\caption{Singular case 1: Histogram of orientation errors.}
\label{fig:robusto1}
\end{figure}

\begin{figure}
\center
\includegraphics[width=\linewidth]{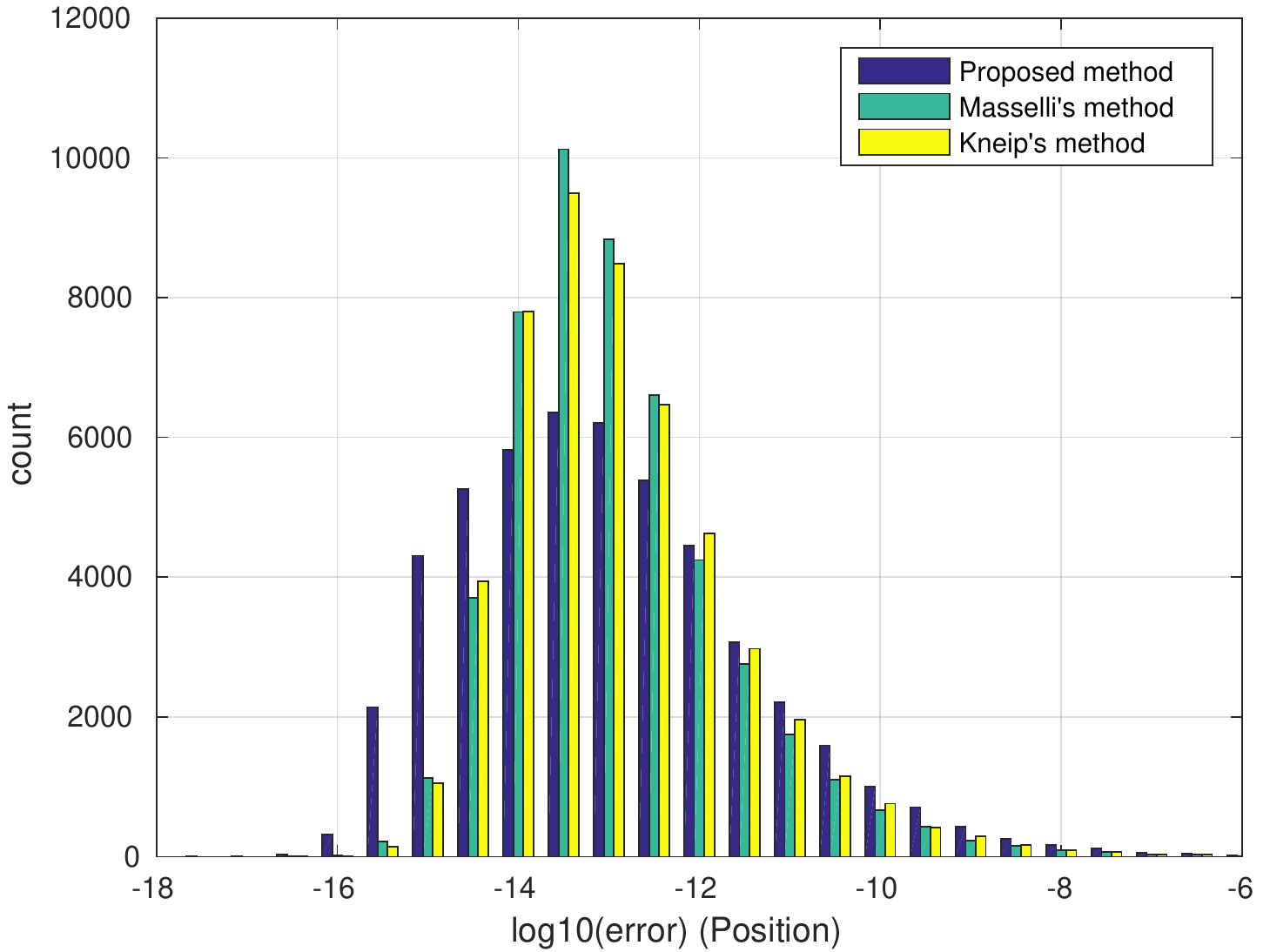}
\caption{Singular case 2: Histogram of position errors.}
\label{fig:robustp2}
\end{figure}

\begin{figure}
\center
\includegraphics[width=\linewidth]{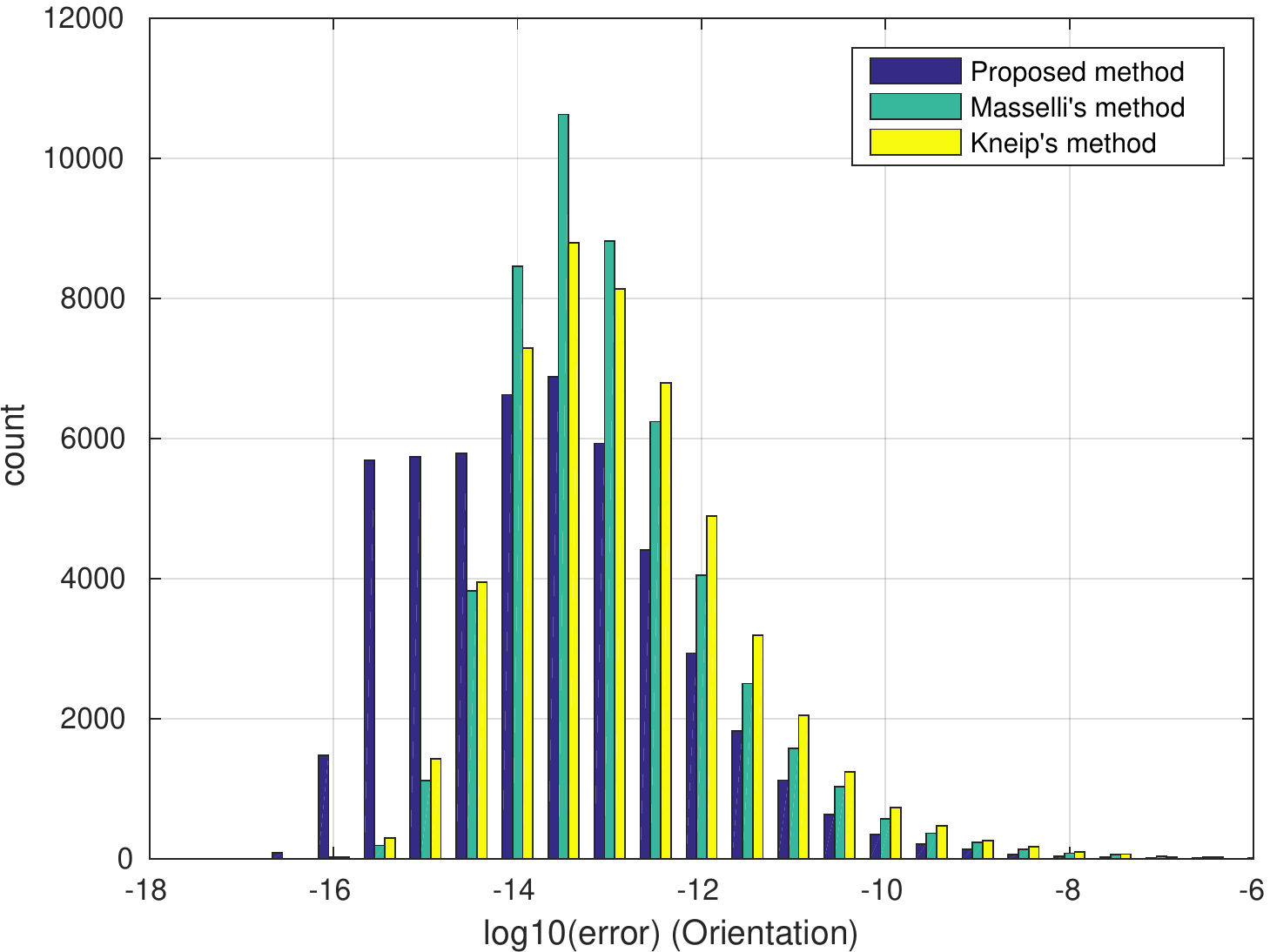}
\caption{Singular case 2: Histogram of orientation errors.}
\label{fig:robusto2}
\end{figure}

There are two typical singular cases that lead to infinite solutions in the P3P problem:
\begin{itemize}
\item Singular case 1: The 3 landmarks are collinear.
\item Singular case 2: Any two of the 3 bearing measurements coincide.
\end{itemize}
In practice, it is almost impossible for these conditions to hold exactly, but we may still have numerical issues when the geometric configuration is close to these cases. 
To test the robustness of the three algorithms considered, we generate simulation data corresponding to small perturbations (uniformly distributed within $[-0.05~~0.05]$) of the features' positions when in singular configurations. 
The errors are defined as in Section~\ref{ssec:numerical}, while we compute the medians of them to assess the robustness of the three methods. 
For fairness, we do not apply root polishing to our code here. 
According to the results shown in Fig.s~\ref{fig:robustp1}~-~\ref{fig:robusto2} and Tables~\ref{tab:robust1}~-~\ref{tab:robust2}, our method achieves the best accuracy in these two close-to-singular cases. The reason is that we do not compute any quantities that may suffer from numerical issues, such as cotangent and tangent in~\cite{kneip2011novel}~and~\cite{masselli2014new}, respectively. 
\begin{table}
\label{tab:robust1}
\center
\begin{tabular}{|c|c|c|}
\hline
 & position & orientation  \\
\hline
Kneip's method & 1.42E-14 & 1.34E-14 \\
\hline
Masselli's method & 7.13E-15 & 6.15E-15\\
\hline
Proposed method & \bf5.16E-15 & \bf3.73E-15 \\
\hline
\end{tabular}
  \caption{Singular case 1: Pose median errors.}
\end{table}
\begin{table}
   \label{tab:robust2}
   
   \center
\begin{tabular}{|c|c|c|}
\hline
 & position & orientation  \\
\hline
Kneip's method & 8.10E-14 & 8.85E-14 \\
\hline
Masselli's method & 7.24E-14 & 6.07E-14\\
\hline
Proposed method & \bf6.73E-14 & \bf1.75E-14 \\
\hline
\end{tabular}
  \caption{Singular case 2: Pose median errors.}
\end{table}
%%%%%%%%%%%%%%%%%%%%%%%%%%%%%%%%%%%%%%%%%%%%%%%%
\section{Conclusion and Future Work}
\label{sec:conclusion}
%%%%%%%%%%%%%%%%%%%%%%%%%%%%%%%%%%%%%%%%%%%%%%%%
In this paper, we have introduced an algebraic approach for computing the solutions of the P3P problem in closed form. 
Similarly to~\cite{kneip2011novel} and \cite{masselli2014new}, our algorithm does \textit{not} solve for the distances first, and hence reduces numerical-error propagation. 
Differently though, it does not involve numerically-unstable functions (\eg, tangent, or cotangent) and has simpler expressions than the two recent alternative methods~\cite{kneip2011novel,masselli2014new}, and thus it outperforms them in terms of speed, accuracy, and robustness to close-to-singular cases.

As part of our ongoing work, we are currently extending our approach to also address the case of the generalized (non-central camera) P3P~\cite{nister2007minimal}. 
\section{Appendix}
\subsection{Proof of $ \mathbf{k}^\prime_3=\mathbf{k}_2\times\mathbf{k}_1$}
\label{ssec:k3p}
First, note that $ \mathbf{k}_2\times\mathbf{k}_1 $ is a unit vector since $ \mathbf{k}_2 $ is perpendicular to $ \mathbf{k}_1 $. Also, from \eqref{eq:k2xk1} and \eqref{eq:t2} we have
\begin{equation}
\mathbf{k}_1^\sT\mathbf{k}^\prime_3=\mathbf{k}_1^\sT\mathbf{C}(\mathbf{k}_2,\theta_2)\mathbf{k}_3=0
\end{equation}
Then, we can prove $ \mathbf{k}^\prime_3=\mathbf{k}_2\times\mathbf{k}_1$ by showing that their inner product is equal to 1:
\begin{align}
(\mathbf{k}_2\times\mathbf{k}_1)^\sT\mathbf{k}_3^\prime&=\mathbf{k}_1^\sT(\mathbf{k}_3^\prime\times\mathbf{k}_2)\nonumber\\
&\stackrel{\eqref{eq:ki}}{=}\frac{\mathbf{k}_1^\sT(\mathbf{k}_3^\prime\times(\mathbf{k}_1\times\mathbf{k}_3))}{\|\mathbf{k}_1\times\mathbf{k}_3\|}\nonumber\\
&\stackrel{\eqref{eq:st2}}{=}\frac{\mathbf{k}_1^\sT(\mathbf{k}_1(\mathbf{k}_3^\sT\mathbf{k}_3^\prime)-\mathbf{k}_3(\mathbf{k}_1^\sT\mathbf{k}_3^\prime))}{\cos\theta_2}\nonumber\\
&=\frac{\mathbf{k}_3^\sT\mathbf{C}(\mathbf{k}_2,\theta_2)\mathbf{k}_3}{\cos\theta_2}=1\nonumber
\end{align}
\subsection{Equivalence between the two solutions of $ \theta_2 $} \label{ssec:theta2}
When solving for $ \theta_2 $ [see \eqref{eq:st2}], we have two possible solutions $ \theta_2^{(1)}=\arccos(\mathbf{k}_1^\sT\mathbf{k}_3)-\frac{\pi}{2} $ and $ \theta_2^{(2)}=\theta_2^{(1)}+\pi $. Next, we will prove that using $ \theta_2^{(2)} $ to find $ {}^\sG_\sCn\mathbf{C} $ is equivalent to using $ \theta_2^{(1)} $.
First, note that (see Fig.~\ref{fig:ki})
\begin{align}
\mathbf{C}(\mathbf{k}_2,\theta_2^{(1)}+\frac{\pi}{2})\mathbf{k}_3&=\mathbf{C}(\mathbf{k}_2,\frac{\pi}{2})\mathbf{k}_3^\prime=-\mathbf{k}_2\times\mathbf{k}_3^\prime\nonumber\\
&=-\mathbf{k}_2\times(\mathbf{k}_2\times\mathbf{k}_1)=\mathbf{k}_1\label{eq:t290}
\end{align}
Then, we can write $ \mathbf{C}(\mathbf{k}_2,\theta_2^{(2)}) $ as
\begin{align}
&\mathbf{C}(\mathbf{k}_2,\theta_2^{(2)})\nonumber\\
=&\mathbf{C}(\mathbf{k}_2,\theta_2^{(1)}+\frac{\pi}{2})\mathbf{C}(\mathbf{k}_2,\frac{\pi}{2})\nonumber\\
=&\mathbf{C}(\mathbf{k}_2,\theta_2^{(1)}+\frac{\pi}{2})\mathbf{C}(\mathbf{k}_3,\pi)\mathbf{C}(\mathbf{k}_2,-\frac{\pi}{2})\mathbf{C}(\mathbf{k}_3,-\pi)\nonumber\\
=&\mathbf{C}(\mathbf{C}(\mathbf{k}_2,\theta_2^{(1)}+\frac{\pi}{2})\mathbf{k}_3,\pi)\mathbf{C}(\mathbf{k}_2,\theta_2^{(1)}+\frac{\pi}{2})\mathbf{C}(\mathbf{k}_2,-\frac{\pi}{2})\mathbf{C}(\mathbf{k}_3,\pi)\nonumber\\
\stackrel{\eqref{eq:t290}}{=}&\mathbf{C}(\mathbf{k}_1,\pi)\mathbf{C}(\mathbf{k}_2,\theta_2^{(1)})\mathbf{C}(\mathbf{k}_3,\pi)\label{eq:ck2}
\end{align}
If we use $ \theta_2^{(2)} $ to find $ {}^\sG_\sCn\mathbf{C} $,
\begin{align}
{}^\sG_\sCn\mathbf{C}&=\mathbf{C}(\mathbf{k}_1,\theta_1^{(2)})\mathbf{C}(\mathbf{k}_2,\theta_2^{(2)})\mathbf{C}(\mathbf{k}_3,\theta_3^{(2)})\nonumber\\
&\stackrel{\eqref{eq:ck2}}{=}\mathbf{C}(\mathbf{k}_1,\theta_1^{(2)})\mathbf{C}(\mathbf{k}_1,\pi)\mathbf{C}(\mathbf{k}_2,\theta_2^{(1)})\mathbf{C}(\mathbf{k}_3,\pi)\mathbf{C}(\mathbf{k}_3,\theta_3^{(2)})\nonumber\\
\label{eq:cccp}&=\mathbf{C}(\mathbf{k}_1,\theta_1^{(2)}+\pi)\mathbf{C}(\mathbf{k}_2,\theta_2^{(1)})\mathbf{C}(\mathbf{k}_3,\theta_3^{(2)}+\pi)
\end{align}
Note that $ {}^\sG_\sCn\mathbf{C} $ in \eqref{eq:cccp} is of the same form as that in \eqref{eq:ccc}, so any solutions of $ {}^\sG_\sCn\mathbf{C} $ computed using $ \theta_2^{(2)} $ will be found by using $ \theta_2^{(1)} $. Thus, we do not need to consider any other solutions for $ \theta_1 $ and $ \theta_3 $ beyond the ones found for $ {}^\sG_\sCn\mathbf{C} $.
\subsection{Determining the sign of $ \sin\theta_1^\prime $}\label{ssec:stheta1p}
From \eqref{eq:st1p}, we have two solutions for $ \sin\theta_1^\prime $, and thus for $\theta_1^\prime $, with $ \theta_1^{\prime(2)}=-\theta_1^{\prime} $. This will also result into two solutions for $ \theta_3^\prime $ [see \eqref{eq:t3}] and, hence, two solutions for $ \theta_3 $: $ \theta_3 $ and $ \theta_3^{(2)}=\theta_3+\pi $. Considering these two options, we get two distinct solutions for $ {}^\sG_\sCn\mathbf{C} $ [see \eqref{eq:cold}]:
\begin{align}
\mathbf{C}_1&\triangleq\mathbf{C}(\mathbf{k}_1,\theta_1^{\prime})\mathbf{C}(\mathbf{k}_3^{\prime\prime},\theta_3)\mathbf{C}(\mathbf{k}_1,\phi)\mathbf{C}(\mathbf{k}_2,\theta_2)\nonumber\\
\mathbf{C}_2&\triangleq\mathbf{C}(\mathbf{k}_1,-\theta_1^{\prime})\mathbf{C}(\mathbf{k}_3^{\prime\prime},\theta_3+\pi)\mathbf{C}(\mathbf{k}_1,\phi)\mathbf{C}(\mathbf{k}_2,\theta_2)\nonumber
\end{align}
Then, notice that
\begin{align}
\mathbf{C}_2\mathbf{C}_1^\sT&=\mathbf{C}(\mathbf{k}_1,-\theta_1^{\prime})\mathbf{C}(\mathbf{k}_3^{\prime\prime},\pi)\mathbf{C}(\mathbf{k}_1,-\theta_1^{\prime})\nonumber\\
&=\mathbf{C}(\mathbf{k}_3^{\prime\prime},\pi)\mathbf{C}(\mathbf{C}^\sT(\mathbf{k}_3^{\prime\prime},\pi)\mathbf{k}_1,-\theta_1^{\prime})\mathbf{C}(\mathbf{k}_1,-\theta_1^{\prime})\nonumber\\
&=\mathbf{C}(\mathbf{k}_3^{\prime\prime},\pi)\mathbf{C}(-\mathbf{k}_1,-\theta_1^{\prime})\mathbf{C}(\mathbf{k}_1,-\theta_1^{\prime})\nonumber\\
&=\mathbf{C}(\mathbf{k}_3^{\prime\prime},\pi)\nonumber
\end{align}
If $ \mathbf{C}_1=\mathbf{C}_2 $, this would require
\begin{align}
\mathbf{C}(\mathbf{k}_3^{\prime\prime},\pi)=\mathbf{C}_2\mathbf{C}_1^\sT=\mathbf{I}\nonumber
\end{align}
which cannot be true, hence $ \mathbf{C}_1 $ and $ \mathbf{C}_2 $ cannot be equal. Thus, there are always two different solutions of $ {}^\sG_\sCn\mathbf{C} $.

If, however, we use the fact that $ d_i\ (i=1,2,3) $ is positive, we can determine the sign of $ \sin\theta_1^\prime $, and choose the valid one among the two solutions of $ {}^\sG_\sCn\mathbf{C} $. Subtracting \eqref{eq:di} pairwise for ($ i=3 $) from ($ i=1 $), we have
\begin{align}
{}^\sG\mathbf{p}_1-{}^\sG\mathbf{p}_3=&d_1{}^\sG_\sCn\mathbf{C}{}^\sCn\mathbf{b}_1-d_3{}^\sG_\sCn\mathbf{C}{}^\sCn\mathbf{b}_3\nonumber\\
 \label{eq:d1-d3}\Rightarrow{}^\sG\mathbf{p}_1-{}^\sG\mathbf{p}_3=&\mathbf{C}(\mathbf{k}_1,\theta_1^{\prime})\mathbf{C}(\mathbf{k}_3^{\prime\prime},\theta_3)\mathbf{C}(\mathbf{k}_1,\phi)\nonumber\\
\cdot&\mathbf{C}(\mathbf{k}_2,\theta_2)(d_1{}^\sCn\mathbf{b}_1-d_3{}^\sCn\mathbf{b}_3)
\end{align}
Multiplying both sides of \eqref{eq:d1-d3} with $ {\mathbf{k}_3^{\prime\prime}}^\sT\mathbf{C}(\mathbf{k}_1,-\theta_1^{\prime}) $ from the left, yields
\begin{align}
&{\mathbf{k}_3^{\prime\prime}}^\sT\mathbf{C}(\mathbf{k}_1,-\theta_1^{\prime})({}^\sG\mathbf{p}_1-{}^\sG\mathbf{p}_3)\nonumber\\
=&{\mathbf{k}_3^{\prime\prime}}^\sT\mathbf{C}(\mathbf{k}_1,\phi)\mathbf{C}(\mathbf{k}_2,\theta_2)(d_1{}^\sCn\mathbf{b}_1-d_3{}^\sCn\mathbf{b}_3)\nonumber\\
\Rightarrow&{\mathbf{k}_3^{\prime\prime}}^\sT(\cos\theta_1^\prime\mathbf{I}+\sin\theta_1^\prime\lfloor\mathbf{k}_1\rfloor+(1-\cos\theta_1^\prime)\mathbf{k}_1\mathbf{k}_1^\sT)\mathbf{u}_1\nonumber\\
=&{\mathbf{k}_3^{\prime}}^\sT\mathbf{C}(\mathbf{k}_2,\theta_2)(d_1{}^\sCn\mathbf{b}_1-d_3{}^\sCn\mathbf{b}_3)\nonumber\\
\stackrel{\eqref{eq:k3pp}}{\Rightarrow}&\sin\theta_1^\prime{\mathbf{k}_3^{\prime\prime}}^\sT\lfloor\mathbf{k}_1\rfloor\mathbf{u}_1=\mathbf{k}_3^\sT(d_1{}^\sCn\mathbf{b}_1-d_3{}^\sCn\mathbf{b}_3)\nonumber\\
\Rightarrow&-\sin\theta_1^\prime\mathbf{u}_1^\sT\lfloor\mathbf{k}_1\rfloor \mathbf{k}_3^{\prime\prime}=-d_3\mathbf{k}_3^\sT{}^\sCn\mathbf{b}_3\nonumber\\
\label{eq:s1pd3}\Rightarrow&\delta\sin\theta_1^\prime=d_3(\mathbf{k}_3^\sT{}^\sCn\mathbf{b}_3)
\end{align}
Using the fact that $ d_3>0 $ and $ \delta>0 $, we select the sign of $ \sin\theta_1^\prime $ to be the same as that of $ \mathbf{k}_3^\sT{}^\sCn\mathbf{b}_3 $. 
\subsection{Least-squares solution for the position}
\label{ssec:ls}
$ {}^\sG\mathbf{p}_\sCn $ can also be solved following a least-squares approach, which is slower but more accurate than \eqref{eq:gpc}. Specifically, \eqref{eq:di} can result in the following system:
\begin{equation}
\begin{bmatrix}
{}^\sG\mathbf{C}_\sCn{}^\sCn\mathbf{b}_1 & & & \mathbf{I}\\
& {}^\sG\mathbf{C}_\sCn{}^\sCn\mathbf{b}_2 & & \mathbf{I}\\
& & {}^\sG\mathbf{C}_\sCn{}^\sCn\mathbf{b}_3 & \mathbf{I}\\
\end{bmatrix}\begin{bmatrix}
d_1\\
d_2\\
d_3\\
{}^\sG\mathbf{p}_\sCn
\end{bmatrix}=\begin{bmatrix}
{}^\sG\mathbf{p}_1 \\
{}^\sG\mathbf{p}_2 \\
{}^\sG\mathbf{p}_3 \\
\end{bmatrix}\nonumber
\end{equation}
Then, we only need to apply QR decomposition~\cite{golub2012matrix} and backsolve for $ {}^\sG\mathbf{p}_\sCn $ (\ie, we do not need to compute $ d_i,\ i=1,2,3 $).

\subsection{Derivation of $ \bar{f}_{ij} $}
\begin{align}
\bar{f}_{i1}&\triangleq\mathbf{u}_i^\sT\lfloor\mathbf{k}_1\rfloor^2\lfloor\mathbf{k}^{\prime\prime}_3\rfloor^2\mathbf{v}^{\prime\prime}_i\nonumber\\
&=\mathbf{u}_i^\sT\lfloor\mathbf{k}_1\rfloor(\mathbf{k}^{\prime\prime}_3\mathbf{k}_1^\sT-(\mathbf{k}_1^\sT\mathbf{k}^{\prime\prime}_3)\mathbf{I})\lfloor\mathbf{k}^{\prime\prime}_3\rfloor\mathbf{v}^{\prime\prime}_i\nonumber\\
&=(\mathbf{u}_i^\sT\lfloor\mathbf{k}_1\rfloor\mathbf{k}^{\prime\prime}_3)(\mathbf{k}_1^\sT\lfloor\mathbf{k}^{\prime\prime}_3\rfloor\mathbf{v}^{\prime\prime}_i)\nonumber\\
&=((\mathbf{u}_i\times\mathbf{k}_1)^\sT\mathbf{k}^{\prime\prime}_3)(\mathbf{k}_1^\sT\mathbf{C}(\mathbf{k}_1,\phi)\mathbf{C}(\mathbf{k}_2,\theta_2)\lfloor\mathbf{k}_3\rfloor\mathbf{v}_i)\nonumber\\
&=\delta{\mathbf{k}_1^\prime}^\sT\lfloor\mathbf{k}_3\rfloor\mathbf{v}_i\nonumber\\
&=\delta\mathbf{v}_i^\sT\mathbf{k}_2\nonumber\\
\bar{f}_{i2}&\triangleq\mathbf{u}_i^\sT\lfloor\mathbf{k}_1\rfloor^2\lfloor\mathbf{k}^{\prime\prime}_3\rfloor\mathbf{v}^{\prime\prime}_i\nonumber\\
&=(\mathbf{u}_i^\sT\lfloor\mathbf{k}_1\rfloor\mathbf{k}^{\prime\prime}_3)(\mathbf{k}_1^\sT\mathbf{v}^{\prime\prime}_i)\nonumber\\
&=\delta(\mathbf{k}_1^\sT\mathbf{C}(\mathbf{k}_1,\phi)\mathbf{C}(\mathbf{k}_2,\theta_2)\mathbf{v}_i)\nonumber\\
&=\delta\mathbf{v}_i^\sT\mathbf{k}_1^\prime\nonumber\\
\bar{f}_{i3}&\triangleq({\mathbf{k}^{\prime\prime}_3}^\sT\mathbf{v}^{\prime\prime}_i)\mathbf{u}_i^\sT\lfloor\mathbf{k}_1\rfloor\mathbf{k}^{\prime\prime}_3\nonumber\\
&=\delta\mathbf{k}_3^\sT\mathbf{C}(\mathbf{k}_2,-\theta_2)\mathbf{C}(\mathbf{k}_1,-\phi)\mathbf{C}(\mathbf{k}_1,\phi)\mathbf{C}(\mathbf{k}_2,\theta_2)\mathbf{v}_i\nonumber\\
&=\delta\mathbf{v}_i^\sT\mathbf{k}_3\nonumber\\
\bar{f}_{i4}&\triangleq-(\mathbf{u}_i^\sT\mathbf{k}_1)\mathbf{k}_1^\sT\lfloor\mathbf{k}^{\prime\prime}_3\rfloor^2\mathbf{v}^{\prime\prime}_i\nonumber\\
&=-(\mathbf{u}_i^\sT\mathbf{k}_1)\mathbf{k}_1(\mathbf{k}^{\prime\prime}_3{\mathbf{k}^{\prime\prime}_3}^\sT-\mathbf{I})\mathbf{v}^{\prime\prime}_i\nonumber\\
&=(\mathbf{u}_i^\sT\mathbf{k}_1)(\mathbf{k}_1\mathbf{v}^{\prime\prime}_i)\nonumber\\
&=(\mathbf{u}_i^\sT\mathbf{k}_1)(\mathbf{v}_i^\sT\mathbf{k}_1^\prime)\nonumber\\
\bar{f}_{i5}&\triangleq-(\mathbf{u}_i^\sT\mathbf{k}_1)\mathbf{k}_1^\sT\lfloor\mathbf{k}^{\prime\prime}_3\rfloor\mathbf{v}^{\prime\prime}_i=-(\mathbf{u}_i^\sT\mathbf{k}_1)(\mathbf{v}_i^\sT\mathbf{k}_2)\nonumber
\end{align}
\subsection{Derivation of $ f_{ij} $ and $ \bar{\bar{\mathbf{C}}} $}
First, note that
\begin{align}
\mathbf{v}_i^\sT\mathbf{k}_2&=({}^\sCn\mathbf{b}_i\times{}^\sCn\mathbf{b}_3)^\sT(\mathbf{k}_1^\prime\times\mathbf{k}_3)\nonumber\\
&=({}^\sCn\mathbf{b}_i^\sT\mathbf{k}_1^\prime)({}^\sCn\mathbf{b}_3^\sT\mathbf{k}_3)-({}^\sCn\mathbf{b}_i^\sT\mathbf{k}_3)({}^\sCn\mathbf{b}_3^\sT\mathbf{k}_1^\prime)\nonumber\\
&=({}^\sCn\mathbf{b}_i^\sT\mathbf{k}_1^\prime)({}^\sCn\mathbf{b}_3^\sT\mathbf{k}_3)\nonumber\\
\mathbf{v}_i^\sT\mathbf{k}_1^\prime&=-({}^\sCn\mathbf{b}_i\times{}^\sCn\mathbf{b}_3)^\sT(\mathbf{k}_2\times\mathbf{k}_3)\nonumber\\
&=-({}^\sCn\mathbf{b}_i^\sT\mathbf{k}_2)({}^\sCn\mathbf{b}_3^\sT\mathbf{k}_3)+({}^\sCn\mathbf{b}_i^\sT\mathbf{k}_3)({}^\sCn\mathbf{b}_3^\sT\mathbf{k}_2)\nonumber\\
&=-({}^\sCn\mathbf{b}_i^\sT\mathbf{k}_2)({}^\sCn\mathbf{b}_3^\sT\mathbf{k}_3)\nonumber
\end{align}
Let $ \psi\triangleq\theta_3-\theta_3^\prime $, and thus
\begin{align}
\label{eq:psi}\begin{bmatrix}
\cos\theta_3^\prime\\
\sin\theta_3^\prime
\end{bmatrix}=\begin{bmatrix}\cos\psi\cos\theta_3+\sin\psi\sin\theta_3 \\ \cos\psi\cos\theta_3+\sin\psi\sin\theta_3\end{bmatrix}
\end{align}
From \eqref{eq:psi} and \eqref{eq:t3p}, we get
\begin{align}
\cos\psi&=\frac{\bar{f}_{11}}{\sqrt{\bar{f}_{11}^2+\bar{f}_{12}^2}}=-\frac{\bar{f}_{15}}{\sqrt{\bar{f}_{14}^2+\bar{f}_{15}^2}}\nonumber\\
&=\frac{\mathbf{v}_1^\sT\mathbf{k}_2}{\sqrt{(\mathbf{v}_1^\sT\mathbf{k}_2)^2+(\mathbf{v}_1^\sT\mathbf{k}_1^\prime)^2}}\nonumber\\
&=\frac{{}^\sCn\mathbf{b}_1^\sT\mathbf{k}_1^\prime}{\sqrt{({}^\sCn\mathbf{b}_1^\sT\mathbf{k}_2)^2+({}^\sCn\mathbf{b}_1^\sT\mathbf{k}_1^\prime)^2}}\nonumber\\
&=\frac{{}^\sCn\mathbf{b}_1^\sT\mathbf{k}_1^\prime}{\|\mathbf{k}_3\times{}^\sCn\mathbf{b}_1\|}\nonumber\\
&={}^\sCn\mathbf{b}_1^\sT\mathbf{k}_1^\prime\nonumber
\end{align}
\begin{align}
\sin\psi&=\frac{\bar{f}_{12}}{\sqrt{\bar{f}_{11}^2+\bar{f}_{12}^2}}=\frac{\bar{f}_{14}}{\sqrt{\bar{f}_{14}^2+\bar{f}_{15}^2}}\nonumber\\
&=\frac{\mathbf{v}_1^\sT\mathbf{k}_1^\prime}{\sqrt{(\mathbf{v}_1^\sT\mathbf{k}_2)^2+(\mathbf{v}_1^\sT\mathbf{k}_1^\prime)^2}}\nonumber\\
%&=-\frac{{}^\sCn\mathbf{b}_1^\sT\mathbf{k}_2}{\sqrt{({}^\sCn\mathbf{b}_1^\sT\mathbf{k}_2)^2+({}^\sCn\mathbf{b}_1^\sT\mathbf{k}_1^\prime)^2}}\nonumber\\
%&=-\frac{{}^\sCn\mathbf{b}_1^\sT\mathbf{k}_2}{\|\mathbf{k}_3\times{}^\sCn\mathbf{b}_1\|}\nonumber\\
&=-{}^\sCn\mathbf{b}_1^\sT\mathbf{k}_2\nonumber
\end{align}
Then, from \eqref{eq:Fi} and \eqref{eq:fs1}, we derive the expressions of $ f_{ij} $:
\begin{align}
f_{11} &= \bar{f}_{11}\cos\psi+\bar{f}_{12}\sin\psi\nonumber\\
&= \delta({}^\sCn\mathbf{b}_3^\sT\mathbf{k}_3)(({}^\sCn\mathbf{b}_1^\sT\mathbf{k}_2)^2+({}^\sCn\mathbf{b}_1^\sT\mathbf{k}_1^\prime)^2)\nonumber\\
&=\delta({}^\sCn\mathbf{b}_3^\sT\mathbf{k}_3)\nonumber\\
f_{21} &= \bar{f}_{21}\cos\psi+\bar{f}_{22}\sin\psi\nonumber\\
&= \delta({}^\sCn\mathbf{b}_3^\sT\mathbf{k}_3)(({}^\sCn\mathbf{b}_2^\sT\mathbf{k}_2)({}^\sCn\mathbf{b}_1^\sT\mathbf{k}_2)+({}^\sCn\mathbf{b}_2^\sT\mathbf{k}_1^\prime)({}^\sCn\mathbf{b}_1^\sT\mathbf{k}_1^\prime))\nonumber\\
&= \delta({}^\sCn\mathbf{b}_3^\sT\mathbf{k}_3)({}^\sCn\mathbf{b}_2^\sT(\mathbf{k}_2\mathbf{k}_2^\sT+\mathbf{k}_1^\prime{\mathbf{k}_1^\prime}^\sT){}^\sCn\mathbf{b}_1)\nonumber\\
&= \delta({}^\sCn\mathbf{b}_3^\sT\mathbf{k}_3)({}^\sCn\mathbf{b}_2^\sT(\mathbf{I}-\mathbf{k}_3\mathbf{k}_3^\sT){}^\sCn\mathbf{b}_1)\nonumber\\
&=\delta({}^\sCn\mathbf{b}_3^\sT\mathbf{k}_3)({}^\sCn\mathbf{b}_2^\sT{}^\sCn\mathbf{b}_1)\nonumber\\
f_{22} &=-\bar{f}_{21}\sin\psi+\bar{f}_{22}\cos\psi\nonumber\\
&= \delta({}^\sCn\mathbf{b}_3^\sT\mathbf{k}_3)(({}^\sCn\mathbf{b}_2^\sT\mathbf{k}_1^\prime)({}^\sCn\mathbf{b}_1^\sT\mathbf{k}_2)-({}^\sCn\mathbf{b}_2^\sT\mathbf{k}_2)({}^\sCn\mathbf{b}_1^\sT\mathbf{k}_1^\prime))\nonumber\\
&= \delta({}^\sCn\mathbf{b}_3^\sT\mathbf{k}_3)({}^\sCn\mathbf{b}_2^\sT(\mathbf{k}_2\mathbf{k}_2^\sT+\mathbf{k}_1^\prime{\mathbf{k}_1^\prime}^\sT){}^\sCn\mathbf{b}_1)\nonumber\\
&= \delta({}^\sCn\mathbf{b}_3^\sT\mathbf{k}_3)({}^\sCn\mathbf{b}_2\times{}^\sCn\mathbf{b}_1)^\sT(\mathbf{k}_1^\prime\times\mathbf{k}_2)\nonumber\\
&= \delta({}^\sCn\mathbf{b}_3^\sT\mathbf{k}_3)\|{}^\sCn\mathbf{b}_2\times{}^\sCn\mathbf{b}_1\|\mathbf{k}_3^\sT\mathbf{k}_3\nonumber\\
&=\delta({}^\sCn\mathbf{b}_3^\sT\mathbf{k}_3)\|{}^\sCn\mathbf{b}_2\times{}^\sCn\mathbf{b}_1\|\nonumber\\
f_{15}&=\bar{f}_{15}\cos\psi-\bar{f}_{14}\sin\psi\nonumber\\
&=-(\mathbf{u}_1^\sT\mathbf{k}_1)f_{11}/\delta\nonumber\\
&=-(\mathbf{u}_1^\sT\mathbf{k}_1)({}^\sCn\mathbf{b}_3^\sT\mathbf{k}_3)\nonumber\\
f_{24}&=\bar{f}_{25}\sin\psi+\bar{f}_{24}\cos\psi\nonumber\\
&=(\mathbf{u}_2^\sT\mathbf{k}_1)f_{22}/\delta\nonumber\\
&=(\mathbf{u}_2^\sT\mathbf{k}_1)({}^\sCn\mathbf{b}_3^\sT\mathbf{k}_3)\|{}^\sCn\mathbf{b}_2\times{}^\sCn\mathbf{b}_1\|\nonumber\\
f_{25}&=\bar{f}_{25}\cos\psi-\bar{f}_{24}\sin\psi\nonumber\\
&=-(\mathbf{u}_2^\sT\mathbf{k}_1)f_{21}/\delta\nonumber\\
&=-(\mathbf{u}_2^\sT\mathbf{k}_1)({}^\sCn\mathbf{b}_3^\sT\mathbf{k}_3)({}^\sCn\mathbf{b}_1^\sT\mathbf{k}_1^\prime)\nonumber
\end{align}
Additionally, we can derive the expression of $ \mathbf{\bar{\bar{C}}} $, which is defined in \eqref{eq:crrc}:
\begin{align}
\mathbf{\bar{\bar{C}}}&=\mathbf{C}(\mathbf{e}_2,\theta_3-\theta_3^\prime)\mathbf{\bar{C}}^\sT\mathbf{C}(\mathbf{k}_1,\phi)\mathbf{C}(\mathbf{k}_2,\theta_2)\nonumber\\
&=\mathbf{C}(\mathbf{e}_2,\psi)\begin{bmatrix}
\mathbf{k}_1 & \mathbf{k}_3^\prime & \mathbf{k}_2
\end{bmatrix}^\sT\mathbf{C}(\mathbf{k}_2,\theta_2)\nonumber\\
&=\mathbf{C}(\mathbf{e}_2,\psi)\begin{bmatrix}
\mathbf{k}_1^\prime & \mathbf{k}_3 & \mathbf{k}_2
\end{bmatrix}^\sT\nonumber\\
&=\begin{bmatrix}
\cos\psi\mathbf{k}_1^\prime-\sin\psi\mathbf{k}_2 & \mathbf{k}_3 & \sin\psi\mathbf{k}_1^\prime+\cos\psi\mathbf{k}_2
\end{bmatrix}^\sT\nonumber\\
&=\begin{bmatrix}
(\mathbf{k}_1^\prime{\mathbf{k}_1^\prime}^\sT+\mathbf{k}_2\mathbf{k}_2^\sT){}^\sCn\mathbf{b}_1 & \mathbf{k}_3 & \sin\psi\mathbf{k}_1^\prime+\cos\psi\mathbf{k}_2
\end{bmatrix}^\sT\nonumber\\
&=\begin{bmatrix}
(\mathbf{I}-\mathbf{k}_3\mathbf{k}_3^\sT){}^\sCn\mathbf{b}_1 & \mathbf{k}_3 & \sin\psi\mathbf{k}_1^\prime+\cos\psi\mathbf{k}_2
\end{bmatrix}^\sT\nonumber\\
&=\begin{bmatrix}
{}^\sCn\mathbf{b}_1 & \mathbf{k}_3 & \sin\psi\mathbf{k}_1^\prime+\cos\psi\mathbf{k}_2
\end{bmatrix}^\sT\nonumber\\
&=\begin{bmatrix}
{}^\sCn\mathbf{b}_1 & \mathbf{k}_3 & {}^\sCn\mathbf{b}_1\times\mathbf{k}_3
\end{bmatrix}^\sT\nonumber
\end{align}

\subsection{Comparison with the P3P code in OpenCV}

\begin{figure}
\center
\includegraphics[width=\linewidth]{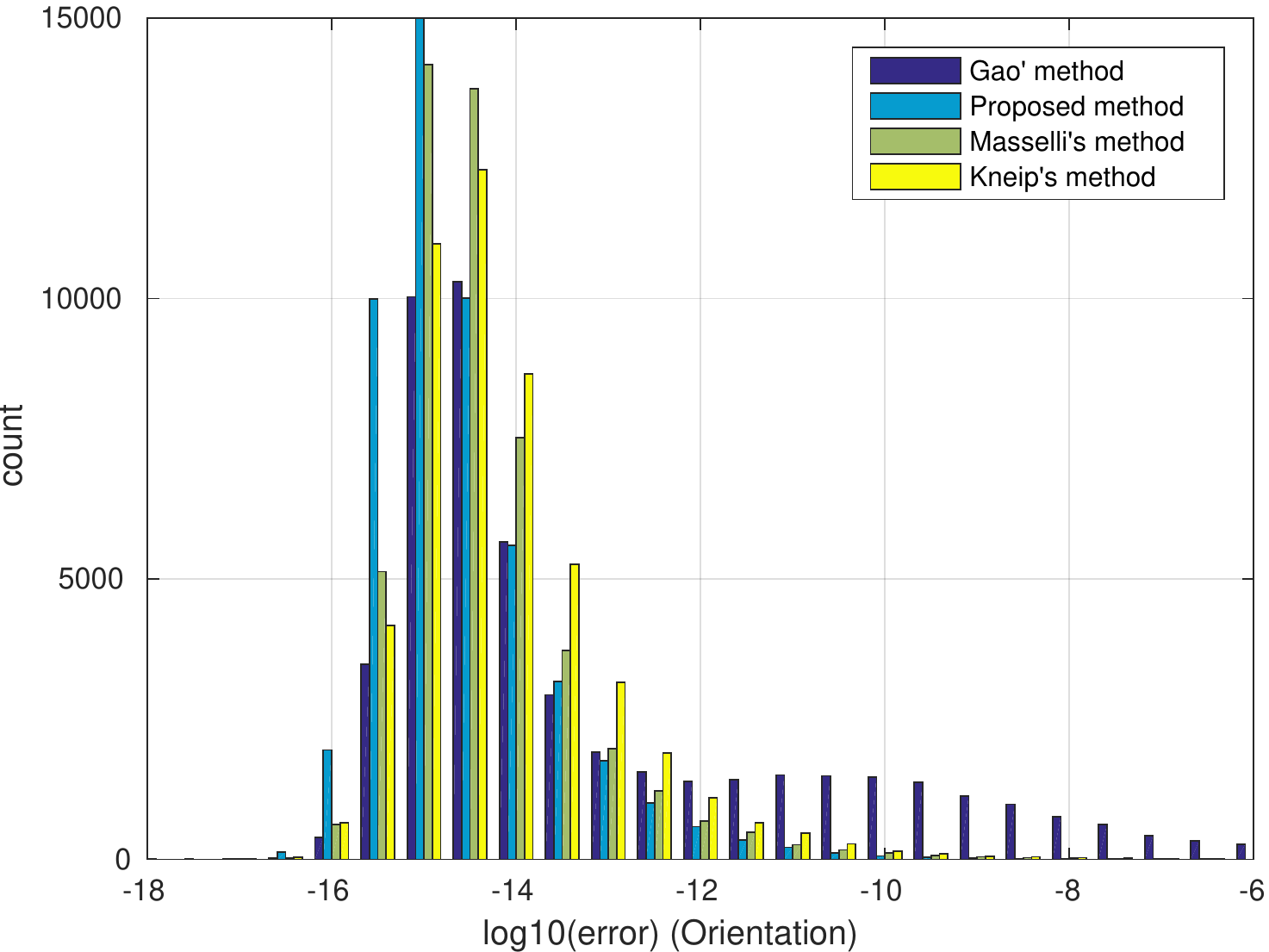}
\caption{Nominal case: Histogram of orientation errors.}
\label{fig:errorOG}
\end{figure}

\begin{figure}
\center
\includegraphics[width=\linewidth]{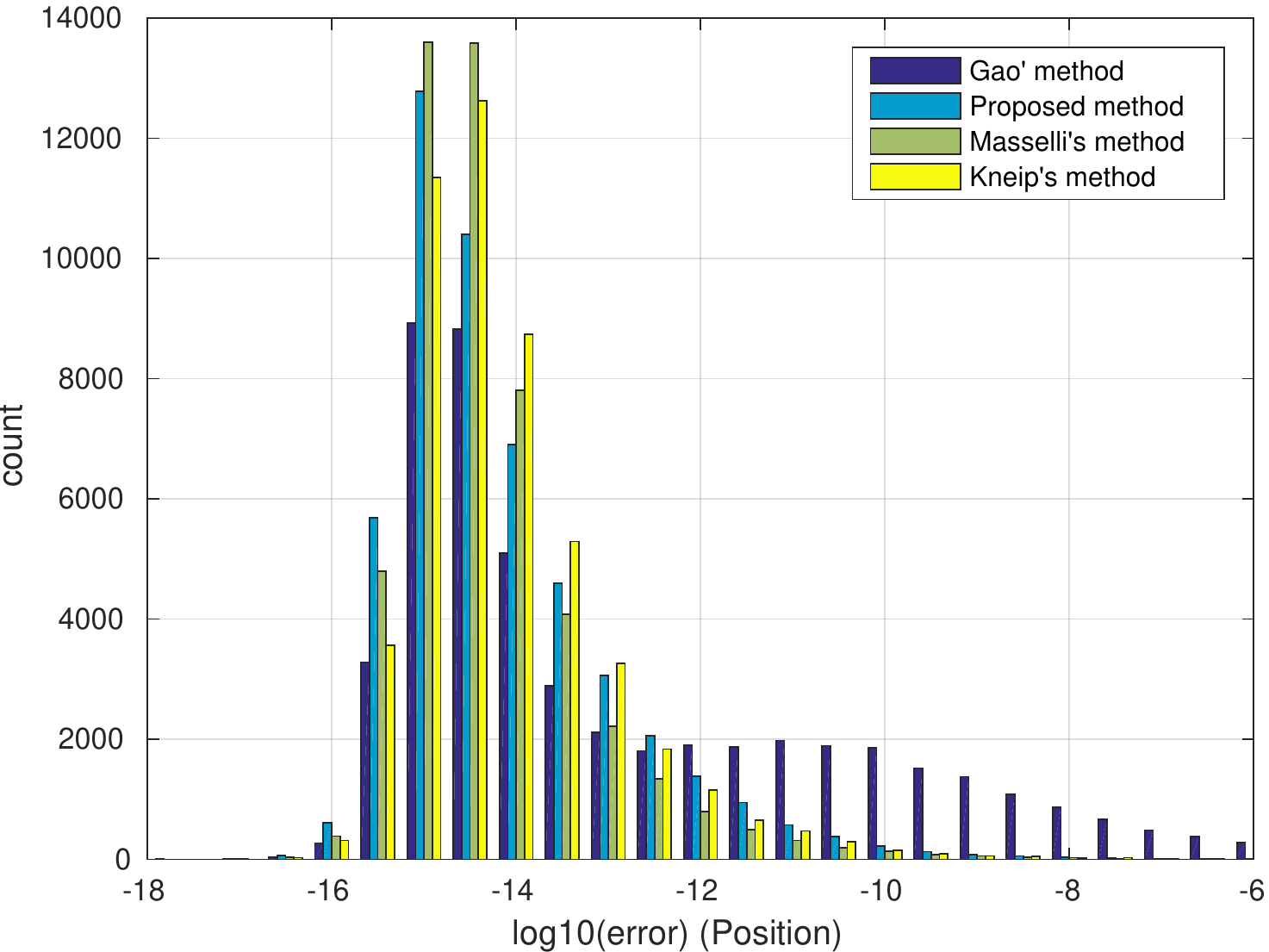}
\caption{Nominal case: Histogram of position errors.}
\label{fig:errorPG}
\end{figure}

We also compared the performance of our code with that in OpenCV (based on \cite{gao2003complete}), using the same setup as Sec.~\ref{sec:results}. The error distributions are showed in Fig.~\ref{fig:errorOG} and Fig.~\ref{fig:errorPG}. It is obvious that the code in OpenCV has lower numerical accuracy comparing to ours. Also, it takes around 3 $ \mu $s on average to compute P3P once, which is much slower than ours (0.52 $ \mu $s according to Sec.~\ref{sec:results}).

In conclusion, our code performs much better than the one in OpenCV.
{
\bibliographystyle{ieee}
\bibliography{references}}
\end{document}